\documentclass[acmsmall,nonacm]{acmart}
\AtBeginDocument{%
  }

\usepackage{algorithmic}
\usepackage{graphicx}
\usepackage{booktabs}
\usepackage{textcomp}
\usepackage{xcolor}
\usepackage{subcaption}
\usepackage{multirow}
\usepackage{tikz}
\usepackage[most]{tcolorbox}
\newcommand*\circled[1]{\tikz[baseline=(char.base)]{
            \node[shape=circle,draw,inner sep=1pt] (char) {#1};}}
\def\tsc#1{\csdef{#1}{\textsc{\lowercase{#1}}\xspace}}
\tsc{WGM}
\tsc{QE}
\tsc{EP}
\tsc{PMS}
\tsc{BEC}
\tsc{DE}

\begin{document}

\title{Augmenting Parameter-Efficient Pre-trained Language Models with Large Language Models}

\author{Saurabh Anand}
\email{saurabh.anand5@tcs.com}
\orcid{0009-0009-7140-2262}
\affiliation{%
  \institution{TCS Research}
  \city{Delhi}
  \country{India}
}

\author{Shubham Malaviya}
\email{shubham.malaviya@tcs.com}
\orcid{0000-0002-9615-5693}

\author{Manish Shukla}
\email{mani.shukla@tcs.com}
\orcid{0000-0003-4867-3530}
\authornote{Corresponding author.}

\author{Sachin Lodha}
\email{sachin.lodha@tcs.com}
\orcid{0000-0001-5771-4977}
\affiliation{%
  \institution{TCS Research}
  \city{Pune}
  \state{Maharashtra}
  \country{India}
}

\renewcommand{\shortauthors}{Anand et al.}

\begin{abstract}
    Training AI models in cybersecurity with help of vast datasets offers significant opportunities to mimic real-world behaviors effectively. However, challenges like data drift and scarcity of labelled data lead to frequent updates of models and the risk of overfitting. To address these challenges, we used parameter-efficient fine-tuning techniques for pre-trained language models wherein we combine compacters with various layer freezing strategies. To enhance the capabilities of these pre-trained language models, in this work we introduce two strategies that use large language models. In the first strategy, we utilize large language models as data-labelling tools wherein they generate labels for unlabeled data. In the second strategy, large language modes are utilized as fallback mechanisms for predictions having low confidence scores. We perform comprehensive experimental analysis on the proposed strategies on different downstream tasks specific to cybersecurity domain. We empirically demonstrate that by combining parameter-efficient pre-trained models with large language models, we can improve the reliability and robustness of models, making them more suitable for real-world cybersecurity applications.
\end{abstract}




\maketitle

\section{Introduction}
The application of artificial intelligence (AI) in cybersecurity has become increasingly critical. This is due to the enormous amount of data generated on daily basis. Most of this data is generated from various cybersecurity applications, such as Security Information and Event Management (SIEM) systems, endpoint security events, and threat intelligence platforms. This data provides rich information for detecting and mitigating cybersecurity-related threats. But it becomes very challenging to analyze such data manually. However, with the development of AI, it is now possible to extract useful insights from these massive datasets (\citet{anandita2023role,kumar2023artificial}).


Various AI models have evolved over the years with BERT-base (\citet{devlin-etal-2019-bert}) to GPT (\citet{brown2020language}) and researchers have harnessed the technique of fine-tuning to fully utilize the power of these models This has resulted in outstanding performance in various downstream tasks such as question answering, sentiment analysis, named entity recognition, etc. (\citet{qiu2020pre}). The application of these models have been extended to cybersecurity domain where they are utilized to handle various cybersecurity related downstream tasks (\citet{ranade2021cybert,aghaei2022securebert,bayer2022cysecbert}). However, the dynamic nature of cybersecurity data creates significant challenges for model fine-tuning and deployment. One critical hurdle is data drift, where models struggle to maintain accuracy over time due to changing data distributions. This evolving nature of data necessitates frequent fine-tuning of models which can incur substantial computational costs. Moreover, there is scarcity of labelled dataset in cybersecurity, which is influenced by legal and ethical constraints, data privacy concerns, resource intensive data labelling process and lack of skilled labours (\citet{cremer2022cyber}). Notably, fine-tuning these millions to billions parameter models on small labelled datasets often lead to the concern of overfitting and performance degradation.

To address the aforementioned issues, researchers have introduced parameter-efficient fine-tuning techniques which keep most of the parameters of pre-trained language models (PLMs) intact and modify only a small subset of parameters. Authors of (\citet{houlsby2019parameter,pfeiffer2020adapterfusion}) have introduced techniques to insert adapters into the architecture of large-scale PLMs. These adapters consist of small trainable layers and during the fine-tuning process only these modules are updated while keeping the most parameters of PLMs unchanged. Extending upon this work, (\citet{karimi2021compacter}), proposed Compacters to reduce the number of trainable parameters in a pre-trained language model. Their method involves inserting low-rank adapters using hypercomplex multiplication (\citet{zhang2021beyond}) into pre-trained models' architecture. Compactors, achieve performance comparable to full fine-tuning on the standard GLUE (\citet{wang2018glue}) and SuperGLUE (\citet{wang2019superglue}) benchmarks. Despite their efficacy, little attention has been given to evaluating the efficiency of compacters in the cybersecurity domain.

In initial version of our work (\citet{saurabh2024sac}), we proposed parameter-efficient pre-trained models, CompFreeze, with unique combination of compacters with various layer freezing strategies to address the above-mentioned issues, which is also covered in this work. In CompFreeze, compacters are inserted into different layers of PLMs while freezing the remaining layers. This results in models which are computationally efficient while maintaining competitive performance with vanilla PLMs. We mainly focus on three different pre-trained models, CyBERT (\citet{ranade2021cybert}), SecureBERT (\citet{aghaei2022securebert}), and CySecBERT (\citet{bayer2022cysecbert}). These models are pre-trained on cybersecurity corpora and are considered as base models for CompFreeze. We compare and evaluate the computational efficiency of these models on different downstream tasks such as Spam Detection, Domain Generation Algorithm (DGA) Classification and entity extraction from Cyber Threat Intelligence sources.  In this work, we have covered CompFreeze-based models and proposed few strategies to enhance them.

CompFreeze-based models offer excellent computational advantages and require access to labelled data. In different cybersecurity applications there is limited access to labelled data. This presents challenges in maximizing the effectiveness of CompFreeze-based models. Additionally, when labeled data representing a particular label is less or a new type of label is introduced, the model may produce predictions with low confidence score. These may affect the performance of the models. In this study, we will investigate how large language models (LLMs) can be utilized along with CompFreeze-based models for more flexible cybersecurity applications.


Large Language Models (LLMs) such as GPT (\citet{brown2020language}), LLaMA (\citet{touvron2023LLaMA}), Gemini (\citet{team2024gemma}), etc. have shown remarkable capabilities in zero-shot learning scenarios for various downstream tasks. They display excellent performance without the need of task-specific fine-tuning and are useful in situations where there is scarcity of labelled datasets. They can quickly adapt to a wide array of tasks with the help of prompt engineering. Although, these LLMs perform well on general domain, they struggle with domain-specific tasks. Additionally, fine-tuning LLMs to handle different downstream tasks can be resource-intensive. To address these challenges, we propose combining domain-specific strengths of CompFreeze-based models and adaptability of LLMs. In this work, we propose two strategies where LLMs can work with CompFreeze-based models: 1) CompFreeze-based models are dependent on small but labelled datasets for effective fine-tuning, but in cybersecurity there is always a scarcity of labelled data. We propose to utilize LLMs to label some volume of unlabeled datasets. These datasets can further be utilized to fine-tune the CompFreeze-based models. 2) While these CompFreeze-based models are computationally efficient and provide remarkable accuracy, there are scenarios where the confidence score for predictions are low. In such cases, we forward the inputs with low confidence scores to LLMs to obtain auxiliary predictions. In summary, we propose a confidence-based mechanism that employs LLMs to provide more reliable inferences for predictions with low confidence scores.

In this work, we aim to combine the efficiency of CompFreeze-based models with the flexible and generalization capabilities of LLMs to produce a more robust application for various cybersecurity downstream tasks. We comprehensively evaluate these strategies on the following three downstream tasks: 1) Spam detection. 2) Domain Generation Algorithm (DGA) classification and, 3) entity extraction from Cyber Threat Intelligence (CTI) sources. These tasks are specific to cybersecurity domain. Our results show that LLMs can be effectively combined with CompFreeze-based models, ultimately leading to more improved and robust cybersecurity application.

Section \ref{sec:related_work} outlines the Related Work. Section \ref{sec:methodology} provides a detailed overview of the methodology employed in this work. In Section \ref{sec:experiments}, we discuss the datasets used for various downstream tasks and the associated experimental results. Section \ref{sec:discusion} includes the discussion of our findings. Finally, Section \ref{sec:conclusion} concludes the paper and identifies a few potential directions for future research.

\section{Related Work}\label{sec:related_work}

\subsection{Parameter-efficient Fine-tuning}
The most common and widely followed approach is to use self-supervised language modeling to pre-train a language model and then fine-tune it for use in many downstream tasks. This widespread paradigm has shown excellent accuracy in a wide variety of NLP tasks, which include single-sentence classification tasks, question-answering tasks, paraphrase tasks, and inference tasks (\citet{devlin-etal-2019-bert,liu2019roberta,conneau-etal-2018-xnli,brown2020language}).  However, such techniques require saving a copy of these large language models and a new set of weights for each task, which becomes quite inefficient. To handle these shortcomings, various parameter-efficient fine-tuning strategies (\citet{li-liang-2021-prefix,houlsby2019parameter,pfeiffer2020adapterfusion,malaviya2023reducing}) have been proposed over the last few years. Adapter-based fine-tuning (\citet{houlsby2019parameter}), where `adapters' are inserted in PLMs, quickly gained popularity in NLP with various applications. Many previous works have shown that it achieves comparable performance as compared to fine-tuning (\citet{wang2020k,pfeiffer2020adapterhub,bapna2019simple}). Following adapters in 2021, \citet{karimi2021compacter} proposed Compacters that utilize the Kronecker Product, low-rank matrices, and parameter sharing across layers to produce adapter weights. However, current research mostly emphasizes the parameter-efficient aspect while ignoring further investigation into its effectiveness. In this paper, we try to bridge this gap and focus on answering the question - `Besides being parameter-efficient, can compacter-based fine-tuning be made more effective?'
There have been some studies that focus on fine-tuning PLMs in low-resource scenarios (\citet{zhang2020revisiting,jiang2019smart,dodge2020fine})). According to earlier research, when working with large-scale parameters, fine-tuning a small number of samples can result in overfitting and poor generalization, leading to less accuracy. \citet{howard2018universal} gradually unfreeze the layers, starting from the last layer, for fine-tuning purposes and to reduce catastrophic forgetting. \citet{clark2019does} analyze BERT's attention layer and observe differences in the information-capturing abilities of the bottom layers and upper layers. \citet{kovaleva2019revealing} have shown the upper layers of BERT mostly change after task-specific fine-tuning. However, none of these works comprehensively examines the effectiveness of odd and even numbered layers of pre-trained language models across multiple downstream tasks.

On the other hand, various parameter-efficient techniques have been introduced in recent years wherein most of the parameters of PLMs are frozen, with only a subset of parameters being modified. Some work in the past has looked into the effectiveness of adapters (\citet{ruckle2020adapterdrop,he2021effectiveness}) for various downstream tasks. \citet{he2021effectiveness} have shown and compared adapter-based tuning with full fine-tuning in low-resource and cross-lingual tasks. \citet{ruckle2020adapterdrop} compared the effectiveness of bottleneck adapters by removing them from lower levels at training and inference time, thus leading to more efficient models. Contrary to their work, in this study we compare the effectiveness of compacters in various pre-trained language models.

\subsection{LLMs in Cybersecurity}
Large language models (LLMs) are advanced AI models that are trained in vast amounts of data. They have shown great zero-shot and few-shot learning capabilities without the need of labelled dataset. They have not only gained popularity in the general domain but also in cybersecurity domain as they aim to facilitate many tasks which previously required manual efforts and interventions. The large language models such as ChatGPT-4 (\citet{achiam2023gpt}), LLaMA models (\citet{touvron2023LLaMA}), etc. have been utilized in various domains, with little applications in cybersecurity domain as well. Code-based LLMs have already become an integral part of many industries. These LLMs have also found applications in various cybersecurity tasks such as vulnerability detection (\citet{ferrag2023securefalcon}), incidence response (\citet{ahmed2023recommending}) and cybersecurity knowledge assistance (\citet{sultana2023towards}). Despite these applications, there remains a need to investigate how LLMs can be integrated with PLMs to enhance the performance of cybersecurity applications. In this work, we aim to bridge this gap by focusing on the integration of large language models (LLMs) with pre-trained language models (PLMs) for a more robust cybersecurity application suitable for real-world scenarios.

\section{Methodology}\label{sec:methodology}


We propose a hybrid combination of compacters and layer freezing to create efficient fine-tuning of the pre-trained models. The motivation of this combination is to provide flexible and efficient architecture by scaling down the number of trainable parameters of PLMs. It leads to a smaller model size, faster convergence of the model, and a smaller number of parameters to fine-tune for any downstream tasks. 
Our method involves inserting compacters into selected layers of the PLM while the remaining layers are frozen, which preserve their weights during fine-tuning for downstream tasks. Importantly, the output from frozen layers remains constant for a given data point, as these layers remain unaltered across subsequent training iterations. In this work, we propose CompFreeze, a compacter-based pre-trained model combined with various layer freezing strategies. We specifically focus on three different pre-trained models in the cybersecurity domain: CyBERT, CySecBERT, and SecureBERT. CyBERT and CySecBERT models have BERT-base as the base model in their architecture, whereas SecureBERT has RoBERTa-base as the base model. Notably, both base models have 12 transformer layers in their foundational architecture.
\begin{figure}[ht]
	\centering
	\includegraphics[width=0.7\columnwidth]{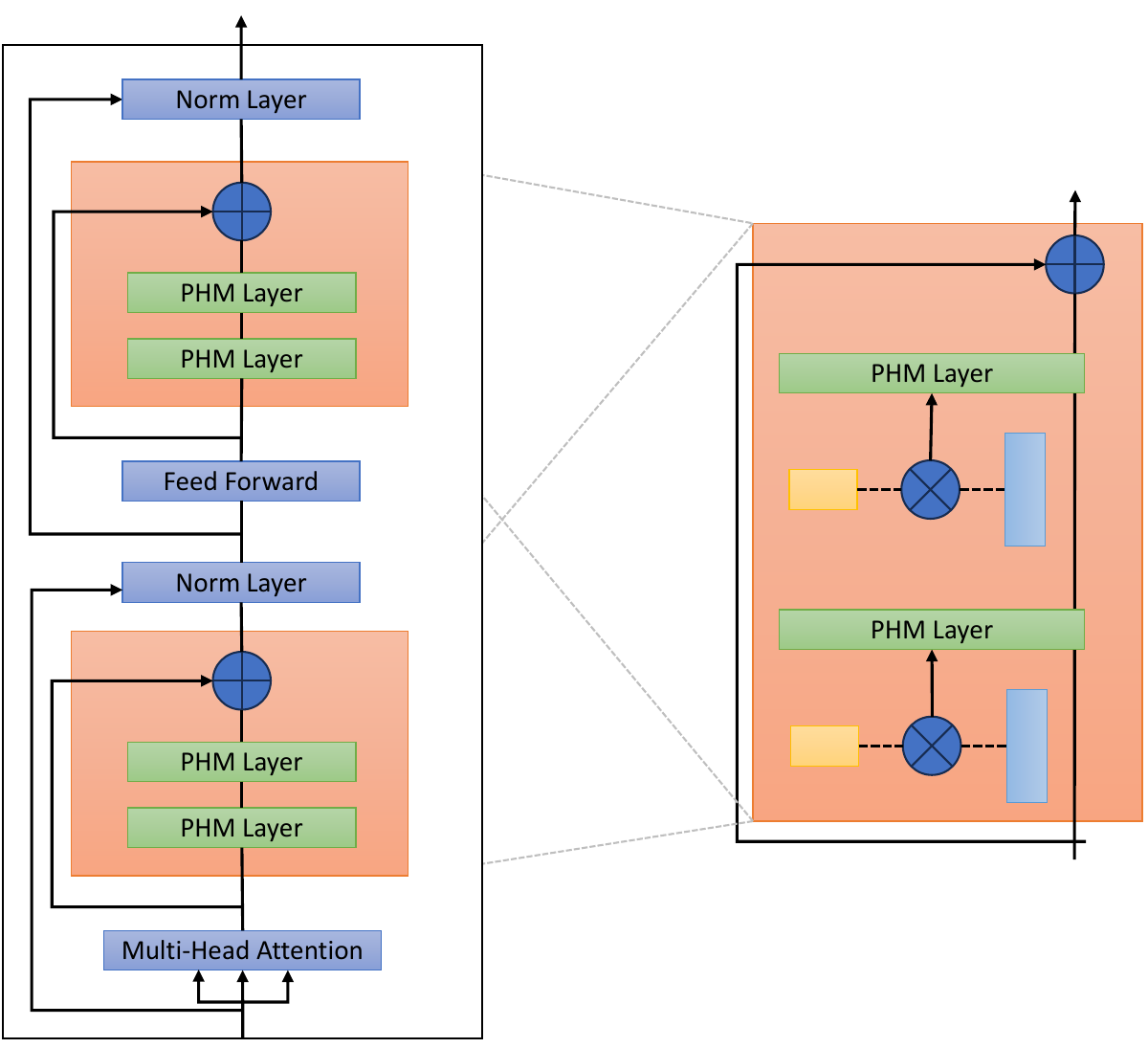}
	\caption{The structure of Compacter adopted from \citet{karimi2021compacter}. }
	\label{fig:compacter}
\end{figure}

\subsection{Compacter-based Tuning}

In the adapter-based tuning approach, a set of lightweight neural networks known as `adapters' is inserted between the PLM transformer layers. These adapters modify only their own parameters for a specific downstream task while keeping the PLM parameters frozen. A notable advantage of this adapter-based tuning method is the creation of a compact model that introduces just a small number of trainable parameters for each task, as opposed to the fine-tuning approach, which necessitates an entirely new model for each task. Extending upon these adapters, \citet{karimi2021compacter} proposed `Compacters'. Figure~\ref{fig:compacter} shows the architecture of Compacter, which generates adapter weights using techniques like the Kronecker product, low-rank matrices, and parameter sharing between layers. A sum of the Kronecker products that corresponds to each parameter W in an adapter is given as follows:
\begin{gather*}
	W = \sum_{i=0}^{n} A_i \otimes B_i \\
	W \in \mathcal{R}^{k \times d}, A_i \in \mathcal{R}^{n \times n}, B_i \in \mathcal{R}^{{\frac{k}{n} \times \frac{d}{n}}}
\end{gather*}
In the equation, $A_i$ and $B_i$ are matrices, $k$ is the size of the input dimension, and $d$ represents the dimension of the compacter. According to (\citet{zhang2021beyond}), a linear layer with this parametrization is known as a parametrized hypercomplex multiplication (PHM) layer. Compacter develops this concept by parameterizing $B_i$ in a manner where all the matrices are of rank at most `r'. All adapter layers share matrix $A_i$ to increase parameter efficiency. The authors of \cite{karimi2021compacter} have shown two different versions of compacters: A single adapter is placed after a feedforward layer (Compacter++), or two adapters are placed per transformer block. These variants are evaluated on Google's T5 model (\citet{raffel2020exploring}).

\begin{figure*}[ht]
	\begin{subfigure}[b]{0.243\textwidth}
		\includegraphics[width=0.9\textwidth]{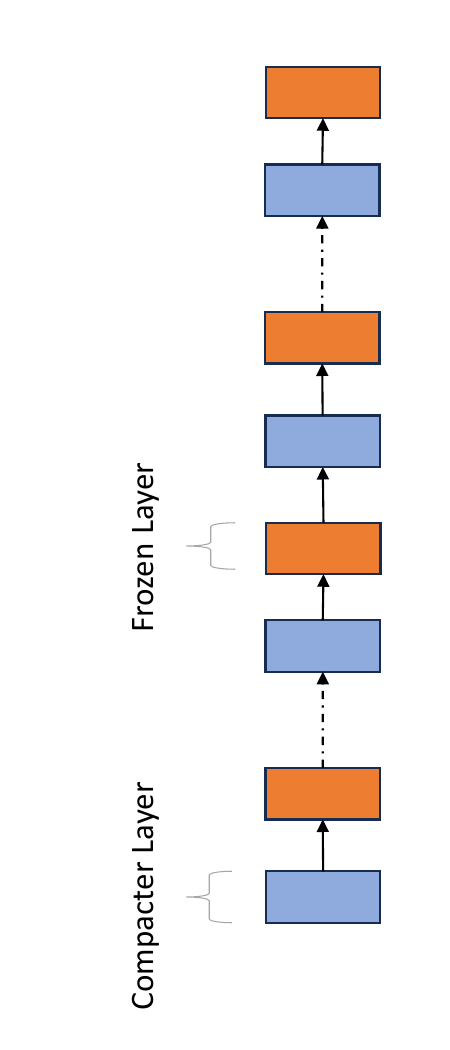}
		\vspace{-0.2cm}
		\caption{Odd-LC} \label{}\vspace{-0.2cm}
	\end{subfigure}
	\begin{subfigure}[b]{0.244\textwidth}
		\includegraphics[width=0.9\textwidth]{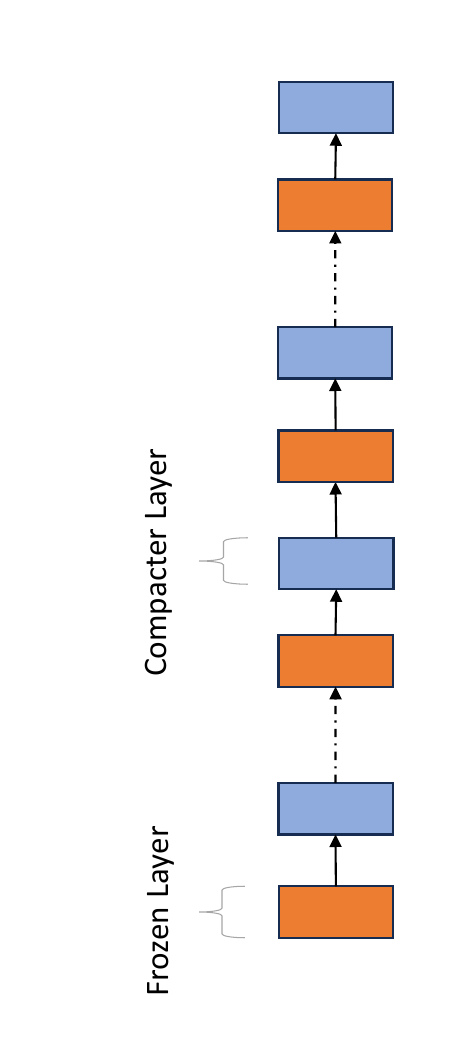}
		\vspace{-0.2cm}
		\caption{Even-LC} \label{}\vspace{-0.2cm}
	\end{subfigure}
	\begin{subfigure}[b]{0.246\textwidth}
		\includegraphics[width=0.9\textwidth]{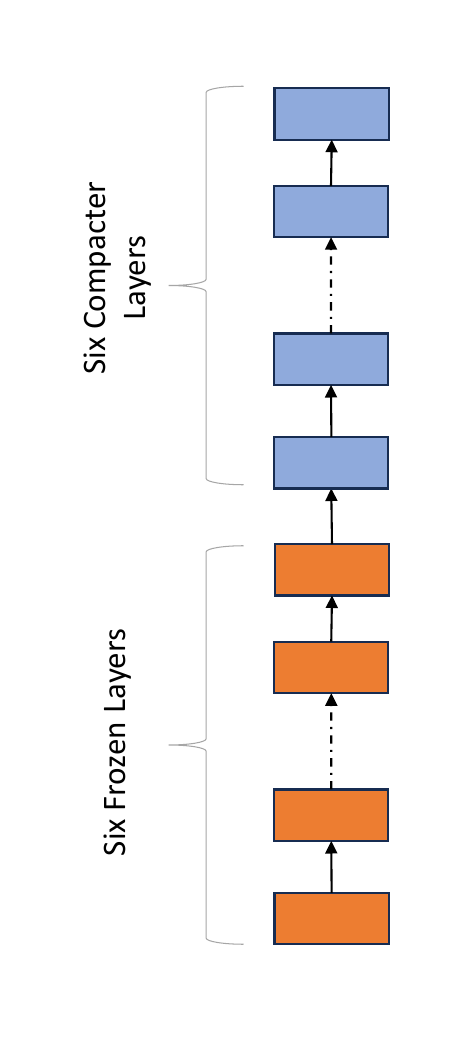}
		\vspace{-0.2cm}
		\caption{Upper-LC} \label{}\vspace{-0.2cm}
	\end{subfigure}
	\begin{subfigure}[b]{0.242\textwidth}
		\includegraphics[width=0.9\textwidth]{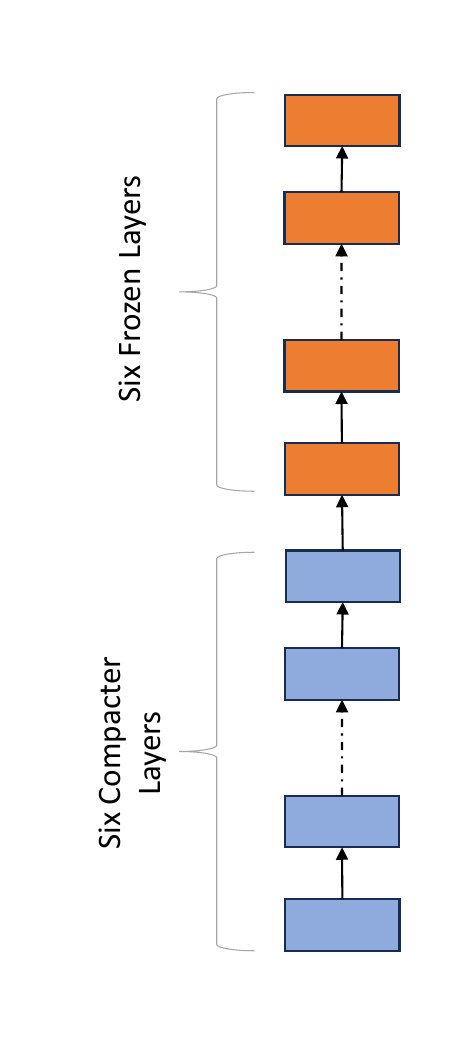}
		\vspace{-0.2cm}
		\caption{Lower-LC} \label{}\vspace{-0.2cm}
	\end{subfigure}
	\caption{Overview of proposed structure. Box in orange are frozen layers of transformers. Box in blue are trainable layers of transformers model in which only compacter parameters are trainable. } 
	\label{fig:Proposed}
	
\end{figure*}

We assess the efficiency of compacter-based PLMs in the field of cybersecurity across diverse downstream tasks, comparing them with the conventional approach of fully fine-tuning vanilla PLMs. Notably, compacter-based PLMs are considerably faster compared to vanilla PLM models, even when trained for the same number of epochs and under identical GPU configurations.
Since most of the parameters are kept frozen while using compacters, backpropagating through a few layers during the training process leads to this speedup.

\subsection{CompFreeze} \label{subsec:CompFreeze_arch}
There has always been a constant demand for effective and sustainable models (\citet{strubell-etal-2019-energy}). By reducing the number of layers involved in the backpropagation step, we can decrease the computational overhead associated with gradient calculations and parameter updates. This leads to faster training times and improves computational efficiency because the backpropagation process involves fewer computations and requires less memory to store intermediate gradients. Additionally, mid layer representations are the most transferable, and top-layer representations are more task-oriented (\citet{merchant-etal-2020-happens}). Furthermore, 50\% of transformers can be safely frozen while retaining the performance accuracy of various downstream tasks (\citet{ingle-etal-2022-investigating}).

Motivated by these observations, we propose CompFreeze, combining compacter with various freezing strategies. These strategies include:

\begin{enumerate}
	\item \textbf{Odd-LC}: Inserting Compacters at Odd numbered transformers layers and freezing the remaining layers. 
	\item \textbf{Even-LC}: Inserting Compacters at Even numbered transformers layers and freezing the remaining layers.
	\item \textbf{Upper-LC}: Inserting Compacters at layer 7 - layer 12 of transformers model and freezing the remaining layers.
	\item \textbf{Lower-LC}: Inserting Compacters at layer 1 - layer 6 of transformers model and freezing the remaining layers.
\end{enumerate}

We study the effectiveness of compacters with given strategies on different cybersecurity-based language models, namely CyBERT, SecureBERT, and CySecBERT. We evaluated these strategies across various cybersecurity-specific downstream tasks. Figure~\ref{fig:Proposed} illustrates the architectural structure of our proposed solution.

\begin{figure*}[ht] 
    \centering
    \begin{subfigure}{0.87\textwidth} 
        \centering
        \includegraphics[width=\textwidth]{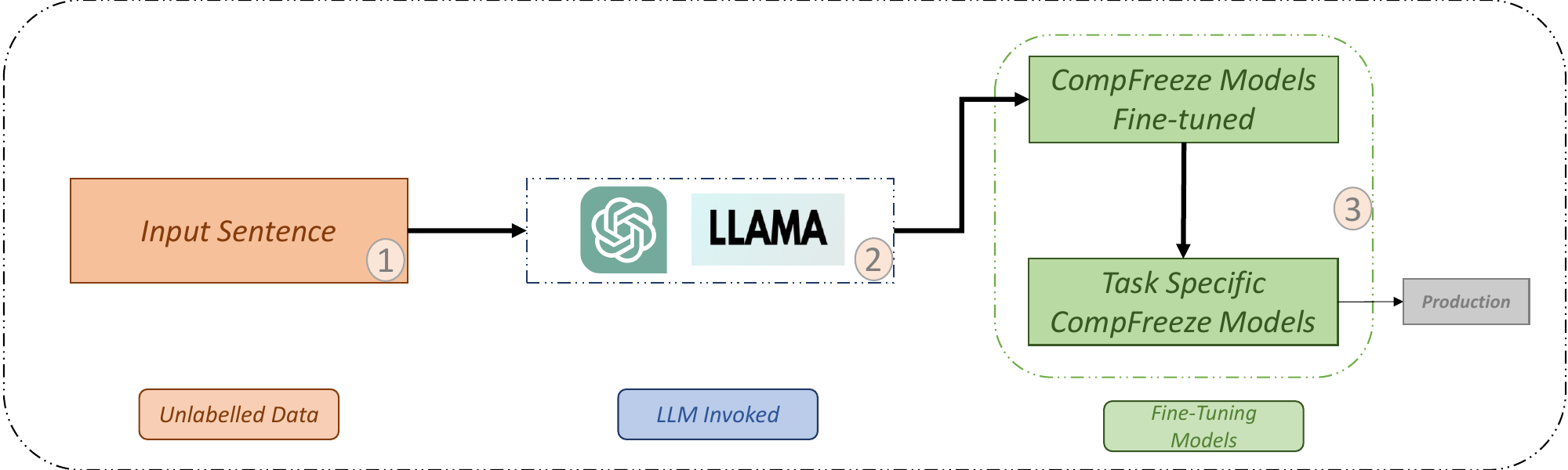} 
        \caption{LLM-assisted Data Labelling for CompFreeze Fine-tuning}
        \label{fig:LLM_data_labelling}
    \end{subfigure}
    \hspace{0.05\textwidth} 
    \begin{subfigure}{0.87\textwidth}
        \centering
        \includegraphics[width=\textwidth]{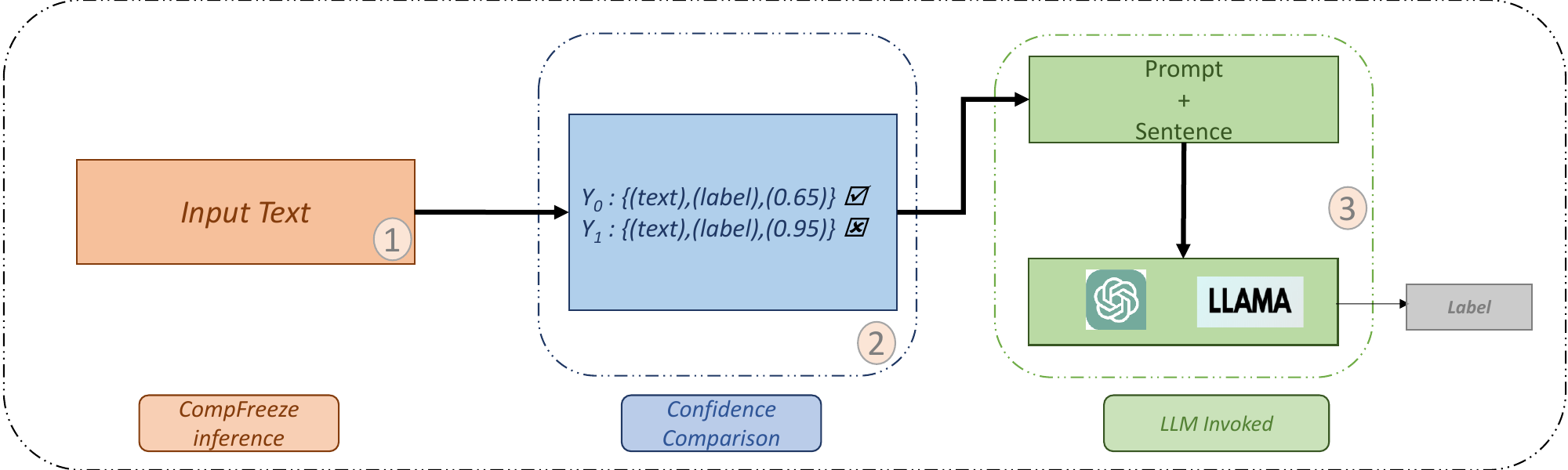} 
        \caption{LLM-based Inference for Low Confidence Prediction} \label{fig:LLM_low_confidence}
    \end{subfigure}
    \caption{Illustration for our LLM Augmented CompFreeze-based models  }
\end{figure*}

\subsection{LLM Augmented CompFreeze-based Models}
CompFreeze-based models provide an excellent alternative to vanilla models in a resource constrained environment. Despite being computationally efficient and having on-par performance, they face limitations in a resource-constrained environment. Their success depends on small but labelled datasets, which is a concern in cybersecurity domain. Furthermore, CompFreeze-based models often struggle with some inputs where they provide low confidence predictions. These prediction diminish the quality of cybersecurity applications. To address these challenges, we propose augmenting CompFreeze-based models with LLMs. In Section \ref{sec:data_labelling} and Section \ref{sec:low_confidence}, we discuss in detail how LLMs can be integrated with CompFreeze-based models to enhance their capabilities. In this work, we will utilize GPT-4 (\citet{achiam2023gpt}) and LLaMA-3 (\citet{dubey2024llama}) versions of LLMs.

\subsubsection{LLM-assisted Data Labelling for CompFreeze Fine-tuning} \label{sec:data_labelling}

In our first proposed strategy, we aim to leverage LLMs to generate labelled data for fine-tuning CompFreeze-based models. This is beneficial as it is quite difficult to get hold of good quality labelled datasets in cybersecurity domain. Under this strategy, we leverage the zero-shot learning abilities of LLMs to label unlabeled data. Figure \ref{fig:LLM_data_labelling}, shows detailed illustration of components involved in this strategy. The workflow of the process is a 3 steps process. At step \circled{1}, a prompt is designed that can provide clear instructions to LLMs. For example, in spam classification task, LLMs should be clearly instructed to classify text as spam or ham labels. Unlabeled data along with prompt is sent to LLM for labelling of data. 
At step \circled{2}, LLM is invoked and it generate labels for unlabeled with the help of prompts provided. At step \circled{3}, the LLM-generated labels are used to fine-tune the CompFreeze-based models. This will allow CompFreeze-based models to adapt to new tasks or data distribution, with minimal performance degradation.

\subsubsection{LLM-based Inference for Low Confidence Prediction} \label{sec:low_confidence}

In our second proposed strategy, we employ LLMs to enhance the prediction quality of low confidence prediction made by CompFreeze-based models. In real world scenarios, models may encounter data outside the scope of their training data. Due to this they may sometimes infer inputs with low confidence. We aim to utilize LLMs as fallback mechanisms for such inferences where CompFreeze-based models are unsure about the prediction. The workflow for this process includes: at step \circled{1}, we calculate the confidence score of input texts for CompFreeze-based model. At step \circled{2}, we define a confidence threshold for particular task. Whenever CompFreeze-based model's prediction is below the defined threshold, that particular prediction is marked as uncertain or ambiguous. Those ambiguous inputs are then sent to LLMs for inference. Finally, at last step \circled{3}, LLMs provide the prediction of those particular inputs along with explanations. These explanations can be analyzed to decide the final prediction for that particular input. We illustrate the detailed components of this strategy in Figure \ref{fig:LLM_low_confidence}.

\section{Experiments}\label{sec:experiments}

\subsection{Datasets}

In this section, we experiment with different downstream tasks that are specific to cybersecurity domain. We compare vanilla PLM with compacter-based PLM and conduct extensive experiments to compare their performance in terms of speed and training time. Statistics of various downstream task datasets are given in Table~\ref{tab:dataset_statistics}.

\textbf{Spam Detection.} This is a text classification task wherein a text is classified as spam or ham. In our study, we use the Enron Email dataset (\citet{klimt2004enron}). The dataset was compiled with approximately 500,000 emails from a total of 150 users. This dataset has been a benchmark dataset for spam detection downstream task. The version used in this work contains a total of 5,857 emails, with 4,361 emails as ham and 1,496 emails as spam.

\textbf{Cyber Threat Intelligence (CTI) Extraction.} In this downstream task, relevant and important tokens are extracted that are useful for studying threat intelligence. \citet{wang2022aptner} collected APT reports from different network security companies and manually labeled those reports for CTI relevant tokens. The dataset contains a total of 21 different categories, with 10,984 sentences, 260,134 tokens, and 39,565 entities in total. The list of entities is given in Table \ref{tab:types}. The final dataset follows BIOES labeling strategy, which includes `B-' tokens that begin a span, `I-' tokens inside a span, `O-' tokens outside of a span, `E-' tokens that end a span, and `S-' tokens with a single span. In this work, we alternatively refer to this dataset as the APTNER dataset.

\textbf{Domain Generation Algorithm Classification.} This downstream task involves identifying domain names that are malicious and those that are benign. This work uses two openly available datasets for this task. The one million Tranco domains (\citet{pochat2018tranco}) are used for benign, non-DGA domains. Tranco is a top-site ranking dataset that is designed for study and is resistant to various manipulations. The majority of previous studies (\citet{lison2017automatic, highnam2021real}) depend on lists of popular sites like the Alexa top one million domains. The Tranco study (\citet{pochat2018tranco}) reveals that it is simple for an opponent to control how these lists are put up. The list of the top one million Alexa users may be changed by enemies with as little as one HTTP request. As a result, the Tranco article develops a million-domain list that is resistant to these manipulations. We use this list as the basis for our DGA domain detector. We only employ 10,000 benign domains out of the 1,000,000 domains on the Tranco list. We utilize the UMUDGA dataset for the DGA malicious domains (\citet{zago2020umudga}). For our work, we randomly included about 10,261 domains from the list. Combined, we have a total of 20,000 domain names, both malicious and benign.

\begin{table}[b]
        \caption{Dataset Statistics}
        \resizebox{0.5\columnwidth}{!}{%
        \begin{tabular}{@{}ccccc@{}}
        \toprule
        \textbf{Task} & \textbf{Description} & \textbf{Train} & \textbf{Test} & \textbf{Label} \\ \midrule
        \textbf{NER} & Cyber Threat Intelligence & 182k & 38k & 21 \\
        \textbf{\begin{tabular}[c]{@{}c@{}}Text\\ Classification\end{tabular}} & Spam Detection & 4k & 1k & 2 \\
        \textbf{\begin{tabular}[c]{@{}c@{}}Text\\ Classification\end{tabular}} & \begin{tabular}[c]{@{}c@{}}Domain Generation Algorithm\\  Classification\end{tabular} & 12k & 4k & 2 \\ \bottomrule
        \end{tabular}%
    }
    \label{tab:dataset_statistics}
\end{table}

\begin{table}[b]
    \caption{Label types in CTI extraction dataset.}
    \resizebox{0.5\columnwidth}{!}{%
    \begin{tabular}{|c|c|}
    \hline
    \textbf{Task} & \textbf{Labels} \\ \hline
    \textbf{\begin{tabular}[c]{@{}c@{}}Cyber Threat\\ Intelligence\end{tabular}} & \begin{tabular}[c]{@{}c@{}}APT, Security Team, Authentication Identity\\ Operating System, Email, Location\\ Time, IP, Domain\\ URL, Protocol, File Names\\ Tool, MD5, SHA1, SHA2\\ Malware, Encryption Algo, Attack Action\\ Vulnerability Names, Vulnerability Number\end{tabular} \\ \hline
    \end{tabular}%
    }
    
    \label{tab:types}
\end{table}

\begin{table*}
\centering
\caption{Percentage of trainable parameters for each task and training time for task. Training time is shown in relative difference with respect to full fine-tuning.}
\resizebox{0.8\columnwidth}{!}{%
		\begin{tabular}{lllll}
		    
			\hline
			\vspace{1pt}

			Task & Model & CyBERT          & SecureBERT     & \multicolumn{1}{c}{CySecBERT} \\ \hline

			&       & Params  \hspace{1pt}   Time & Params \hspace{1pt}   Time & Params  \hspace{1pt}  Time                \\ \cline{3-5}

			CTI Extraction &

			\begin{tabular}[c]{@{}l@{}}Vanilla\\ Odd-LC\\ Even-LC\\ Upper-LC\\ Lower-LC\end{tabular} &

			\begin{tabular}[c]{@{}l@{}}100\%           \hspace{12pt} -\\ 0.11\%     \hspace{1pt} -19.4\%\\ 0.11\%    \hspace{1pt}  -20.1\%\\ 0.11\%     \hspace{1pt} -19.7\%\\ 0.11\%     \hspace{1pt} -19.2\%\end{tabular} &

			\begin{tabular}[c]{@{}l@{}}100\%          \hspace{12pt}  -\\ 0.09\%    \hspace{1pt}  -36.5\%\\ 0.09\%     \hspace{1pt} -40.3\%\\ 0.09\%    \hspace{1pt}  -34.2\%\\ 0.09\%    \hspace{1pt}  -37.6\%\end{tabular}&

			\begin{tabular}[c]{@{}l@{}}100\%          \hspace{12pt} -\\ 0.11\%  \hspace{1pt}     -11.7\%\\ 0.11\%   \hspace{1pt}    -10.6\%\\  0.11\%   \hspace{1pt}    -10.6\%\\  0.11\%    \hspace{1pt}  -11.7\%\end{tabular} \\ \hline

			Spam Detection &

			\begin{tabular}[c]{@{}l@{}}Vanilla\\ Odd-LC\\ Even-LC\\ Upper-LC\\ Lower-LC\end{tabular} &

			\begin{tabular}[c]{@{}l@{}}100\%         \hspace{12pt}  -\\ 0.06\%   \hspace{1pt}   -39.4\% \\ 0.06\%    \hspace{1pt}  -38.8\%\\ 0.06\%   \hspace{1pt}   -39.9\%\\ 0.06\%   \hspace{1pt}   -38.8\%\end{tabular} &

			\begin{tabular}[c]{@{}l@{}}100\%         \hspace{12pt}  -\\ 0.53\%    \hspace{1pt}  -11.6\%\\ 0.53\%    \hspace{1pt}  -14.5\%\\ 0.53\%   \hspace{1pt}   -12.5\%\\ 0.53\%    \hspace{1pt}  -13.6\%\end{tabular} &

			\begin{tabular}[c]{@{}l@{}}100\%         \hspace{12pt}  -\\ 0.06\% \hspace{1pt}     -28.7\%\\ 0.06\%    \hspace{1pt}  -28.4\%\\ 0.06\%   \hspace{1pt}   -28.5\%\\ 0.06\%  \hspace{1pt}    -26.6\%\end{tabular} \\ \hline

			DGA Classification &

			\begin{tabular}[c]{@{}l@{}}Vanilla\\ Odd-LC\\ Even-LC\\ Upper-LC\\ Lower-LC\end{tabular} &

			\begin{tabular}[c]{@{}l@{}}100\%        \hspace{12pt}   -\\ 0.06\%   \hspace{1pt}    -23.6\%\\ 0.06\%    \hspace{1pt}   -22.5\%\\ 0.06\%   \hspace{1pt}    -36.4\%\\ 0.06\%      \hspace{1pt} -27.7\%\end{tabular} &

			\begin{tabular}[c]{@{}l@{}}100\%         \hspace{12pt}  -\\ 0.53\%    \hspace{1pt}   -41.5\%\\ 0.53\%    \hspace{1pt}   -41.2\%\\ 0.53\%  \hspace{1pt}     -40.3\%\\ 0.53\%      \hspace{1pt} -40.4\%\end{tabular} &

			\begin{tabular}[c]{@{}l@{}}100\%        \hspace{12pt}   -\\ 0.06\%   \hspace{1pt}    -32.8\%\\ 0.06\%    \hspace{1pt}   -28.4\%\\ 0.06\%   \hspace{1pt}    -35.6\%\\ 0.06\%      \hspace{1pt} -34.5\%\end{tabular} \\ \hline

		\end{tabular}
	
	}
    
	 \label{tab:parameters}
\end{table*}

\subsection{Implementation Details}

 All the experiments were conducted on NVIDIA MiG A100 GPUs. We use the Opendelta library\footnote{https://opendelta.readthedocs.io/} for initializing the compacter architecture. We use the default reduction factor in compacters as 16. The rank of parameterized hypercomplex multiplication (PHM) layer is used as 1, which is the default value, and the initialization range for PHM is taken as 0.0001. We fine-tune compacter-based PLMs with a range of learning rates, \{1e-5, 2e-5, 3e-5, 4e-5, 5e-5\}, on various downstream tasks. Furthermore, the maximum sequence length for the spam detection task is 512, and it is 128 for all the other downstream tasks. A batch size of 8 is applied for all the experiments. We fine-tune models for three and ten epochs and report the detailed results in Section \ref{sec:results}. We train all models with the AdamW optimizer from the HuggingFace library with default hyper-parameters of $\beta =\{0.9,0.999\},\;\epsilon=1e-8$. We compare the F1 score of CompFreeze-based models on different downstream tasks with the baseline model. Following (\citet{houlsby2019parameter}), we update the layer normalization parameters for all methods. Moreover, we kept the classifier layer parameters for all the models as trainable. For all large language model-based experiments, we utilize the chat version of the models which are publicly available. We compare and evaluate GPT-4 and LLaMA-3 models with CompFreeze-based models on the three downstream tasks specific to cybersecurity domain.
 
 \begin{figure*}[ht]
	\centering
	
	\begin{subfigure}[]{0.32\textwidth}
		\includegraphics[width=1.0\columnwidth]{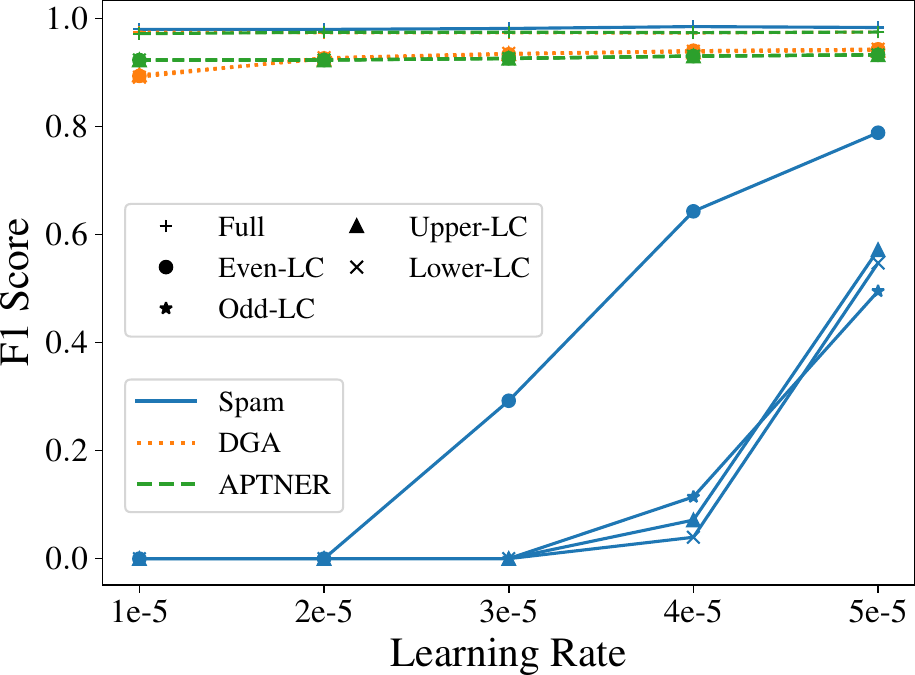}
		\caption{CyBERT, Epochs=3}
		\label{}
	\end{subfigure}
	\begin{subfigure}[]{0.32\textwidth}
		\includegraphics[width=1.0\columnwidth]{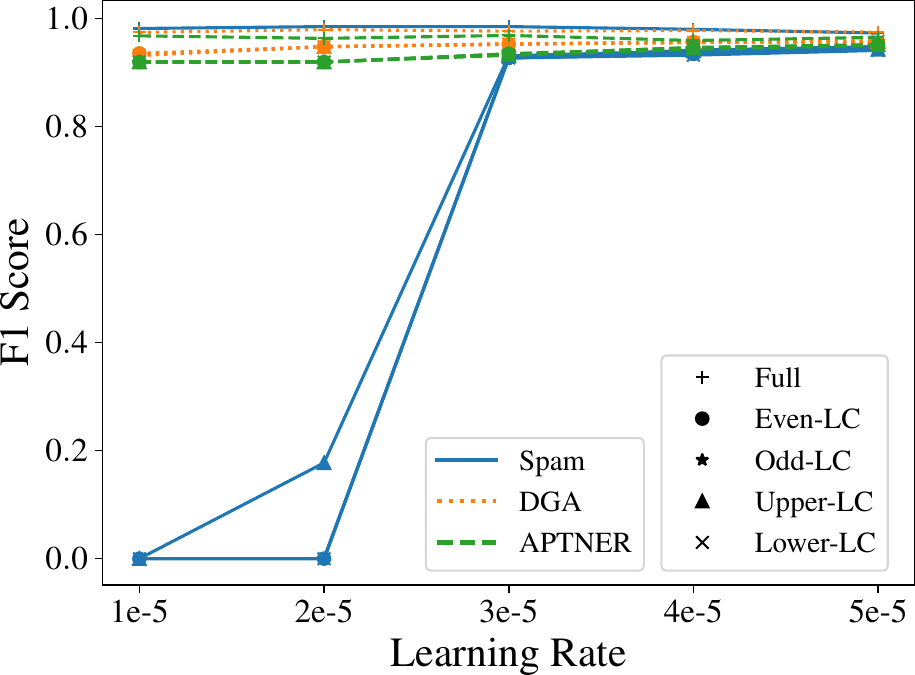}
		\caption{SecureBERT, Epochs=3}
		\label{}
	\end{subfigure}
	\begin{subfigure}[]{0.32\textwidth}
		\includegraphics[width=1.0\columnwidth]{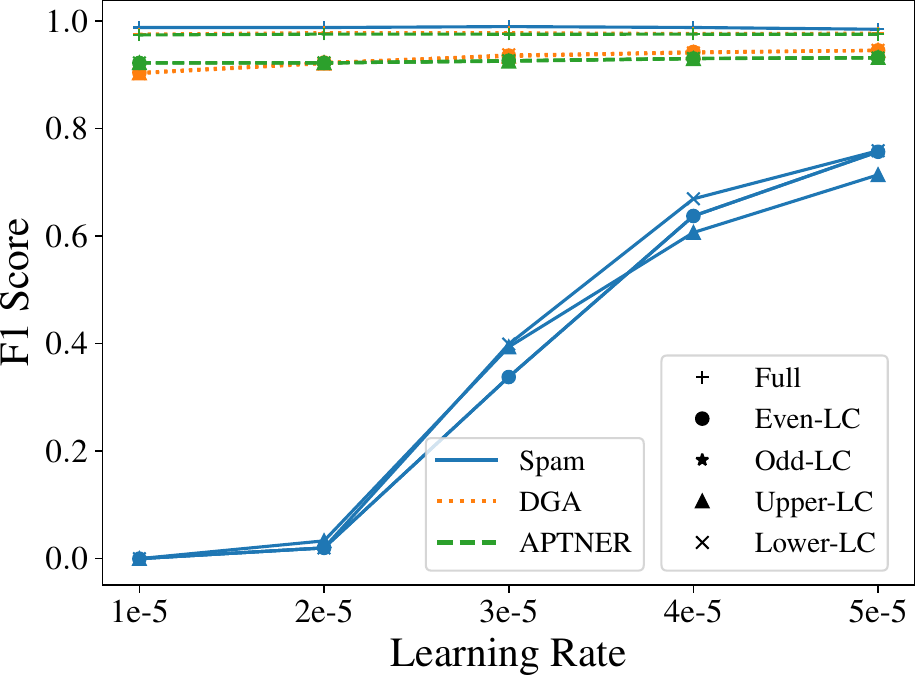}
		\caption{CySecBERT, Epochs=3}
		\label{}
	\end{subfigure}
	
	\begin{subfigure}[]{0.32\textwidth}
		\includegraphics[width=1.0\columnwidth]{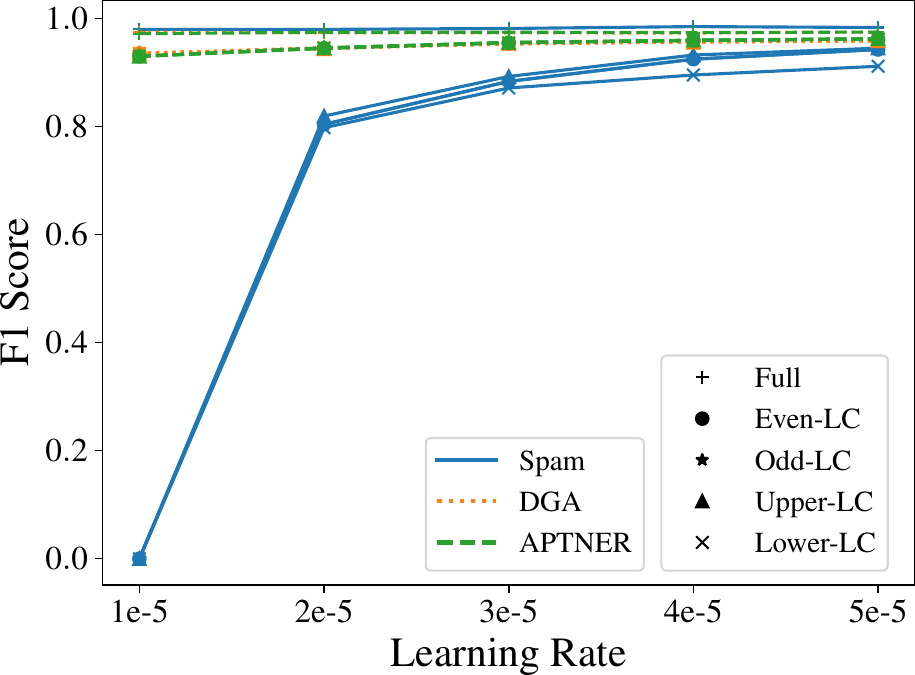}
		\caption{CyBERT, Epochs=10}
		\label{}
	\end{subfigure}
	\begin{subfigure}[]{0.32\textwidth}
		\includegraphics[width=1.0\columnwidth]{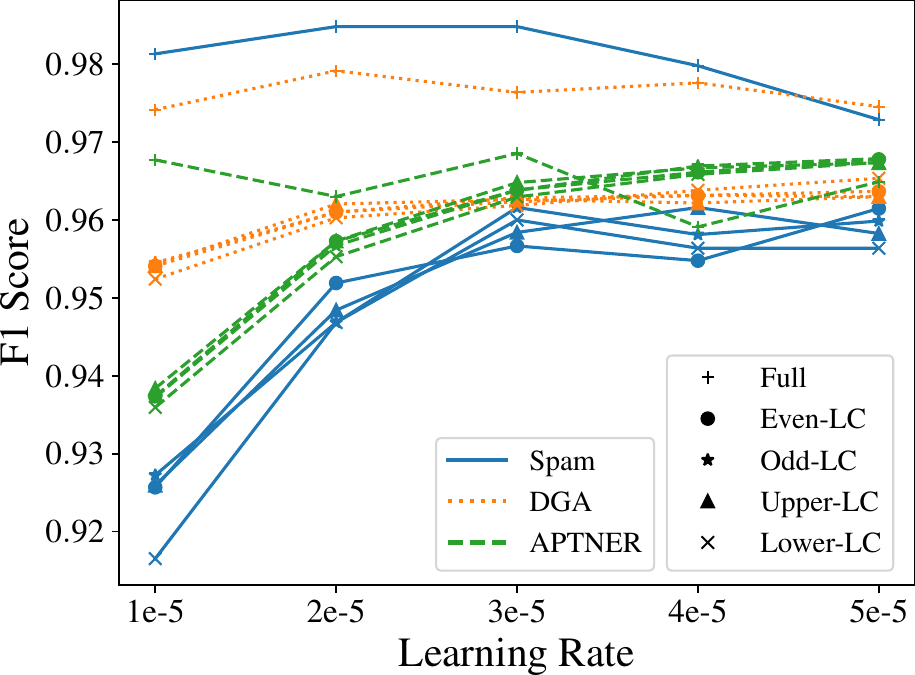}
		\caption{SecureBERT, Epochs=10}
		\label{}
	\end{subfigure}
	\begin{subfigure}[]{0.32\textwidth}
		\includegraphics[width=1.0\columnwidth]{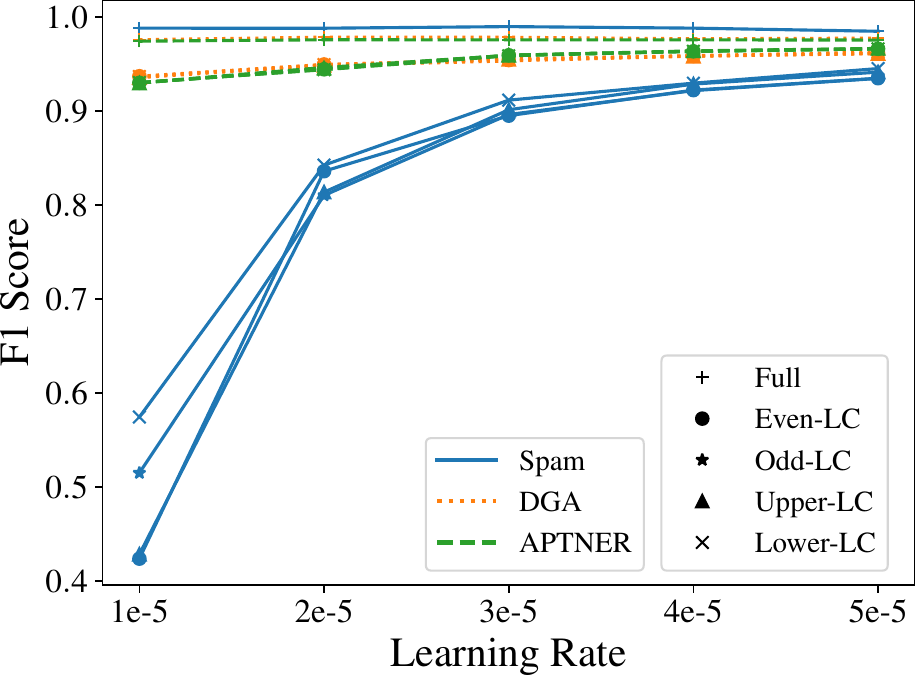}
		\caption{CySecBERT, Epochs=10}
		\label{}
	\end{subfigure}
	
	\caption{Performance comparison for models trained using different learning rates, and epochs 3 and 10.}
	\label{fig:lr_f1}
\end{figure*}

\section{Results}\label{sec:results}
In this section, we present the results of our experiments. Firstly, we present a detailed analysis of various experiments conducted on different CompFreeze-based models. Secondly, we present our results of the two strategies proposed and empirically show that LLMs can be augmented along with CompFreeze-based models for flexible cybersecurity applications.

\subsubsection{Analysis on CompFreeze-based Models }
We evaluate the performance of baseline models with the CompFreeze-based models. We study the efficacy of CompFreeze on datasets of various downstream tasks in the cybersecurity domain.

Table~\ref{tab:parameters} presents a breakdown of training speed and parameter counts for CompFreeze-based models versus baseline models. Our observations show that CompFreeze requires fewer parameters and less training time. It effectively preserves the inherent capabilities of the vanilla model and exhibits minimal performance degradation. CompFreeze consistently achieves results on par with baseline models with different proposed strategies.

\textbf{Learning Rate Analysis:}\label{subsec:lr_analysis}
The learning rate is a critical parameter that influences model training convergence. It also helps to maintain the stability of the training process and impacts the final performance of the model. In this section, we evaluate the effectiveness of our proposed approach when applied to three distinct PLMs adapted to the cybersecurity domain, considering a range of learning rates: \{1e-5, 2e-5, 3e-5, 4e-5, 5e-5\}. In Figure~\ref{fig:lr_f1}, we present a comparative analysis of our proposed method, CompFreeze, with various layer freezing techniques, including Odd-LC, Even-LC, Upper-LC, Lower-LC, and full model fine-tuning, across different datasets. Notably, our proposed method achieves on par performance to full model fine-tuning while employing only 50\% trainable layers. Furthermore, we observed a negligible performance difference among the various layer freezing techniques. All three models yield identical F1-score for the DGA and APTNER datasets, for training epochs three and ten. It is worth noting that the choice of learning rate appears to have a minor impact on model performance for these two datasets. On the other hand, we observe a noticeable impact of learning rate on model performance for the Spam dataset. All three models attain very low F1 scores when trained with learning rates of 1e-5 and 2e-5 for three epochs. Upon inspection, we found that the models were failing to converge within three training epochs. Subsequently, we extended the training to 10 epochs and observed a substantial performance improvement, particularly evident for the Spam dataset. As the learning rate increases from 1e-5 to 5e-5, the model performance also increases consistently. We observed the best performance for a learning rate of 5e-5. This observation aligns with the findings in the BERT paper \citet{devlin-etal-2019-bert}, which recommended a learning rate of 5e-5 for various natural language processing tasks.

From a training convergence perspective, SecureBERT demonstrates faster convergence compared to the other two models, and its performance is relatively less influenced by the choice of learning rate. For instance, when trained for 10 epochs with a learning rate of 1e-5 on the Spam dataset, CyBERT and CySecBERT achieve F1 scores of 0\% and 50\%, respectively, while SecureBERT scores more than 90\%. It is important to note that SecureBERT is based on the RoBERTa architecture, whereas CyBERT and CySecBERT are based on the BERT architecture.

\begin{figure}[ht]
	\centering
	\includegraphics[width=0.5\columnwidth]{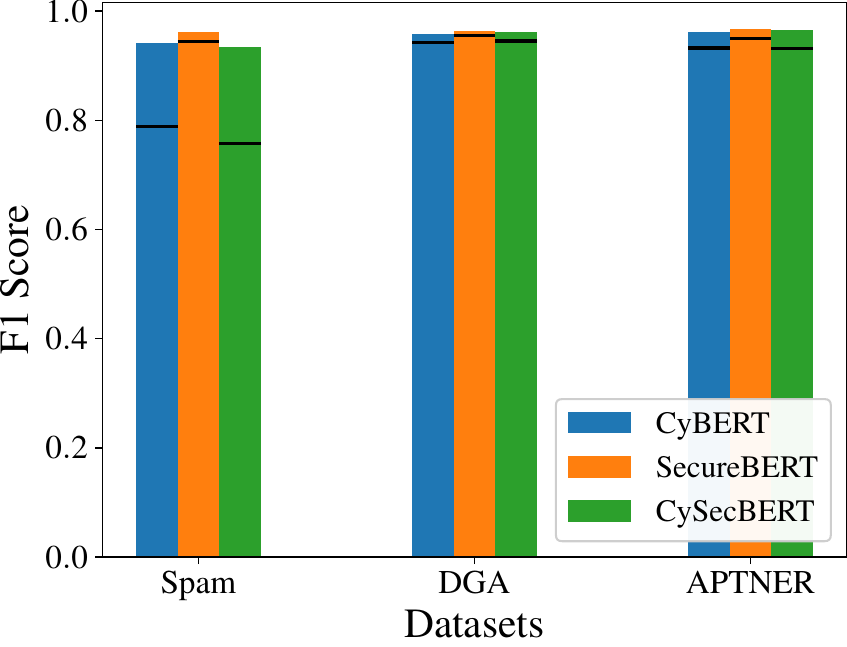}
	\caption{Performance comparison of pre-trained models adopted to cybersecurity domain.}
	\label{fig:model_comparison}
\end{figure}

\textbf{Comparison of Cybersecurity Pre-trained Models} \label{subsec:model_comparison}
In the preceding section, we observed that CyBERT, SecureBERT, and CySecBERT all achieved their highest F1 scores at a learning rate of 5e-5. In this section, we delve into an analysis to determine which model performs optimally across all datasets while constraining the trainable layers to 50\%. Here, we present results for Even-LC technique, although it is important to note that, as discussed earlier, Odd-LC, Upper-LC, and Lower-LC exhibit similar performance to the Even-LC. Figure~\ref{fig:model_comparison}, results are presented for models trained for 10 epochs with a learning rate of 5e-5. Also, the black markers on each bar denote the F1 score when the model is trained for only three epochs.
Notably, for both the DGA and APTNER datasets, all three models yield almost similar performance when trained until convergence, which in our case is 10 epochs. However, it is worth highlighting that, on the Spam dataset, SecureBERT significantly outperforms the other models when they are trained for only three epochs. However, for other datasets, SecureBERT performs slightly better than other models when trained for three epochs. Parameter-efficient fine-tuning is typically employed when data is scarce, and extensive training could lead to model overfitting. In such scenarios, selecting the right model can prove to be advantageous.

\begin{figure}[ht]
	\centering
	\includegraphics[width=0.5\columnwidth]{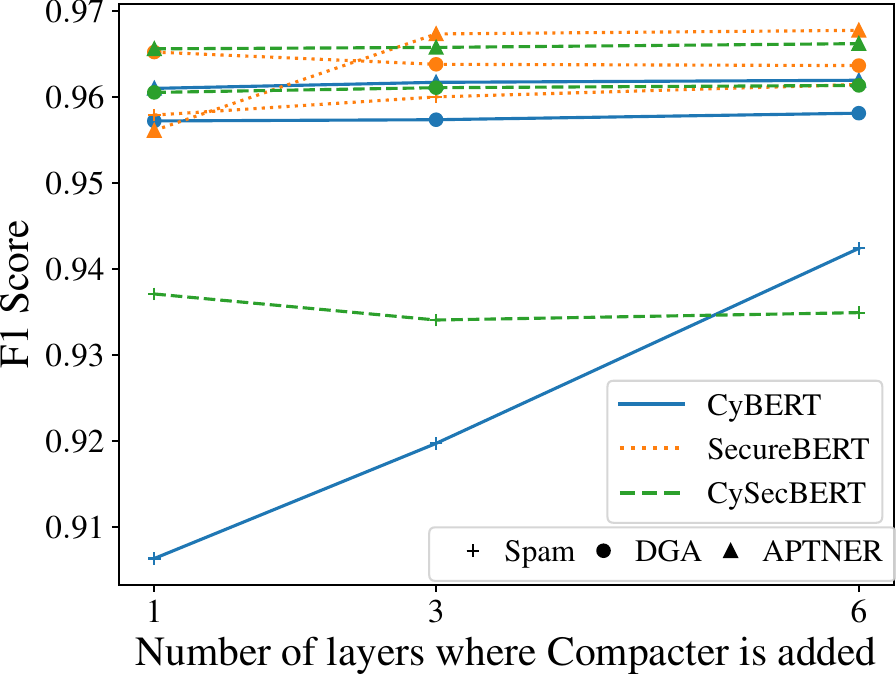}
	\caption{Model performance pertaining to different number of trainable parameters for learning rate=5e-5 and Epochs=10.}
	\label{fig:delta_percent_comparison}
\end{figure}

\textbf{Effect of Varying Number of Compacter Modules}\label{subsec:delta_percentage_comparison}
 As previously demonstrated when employing 50\% trainable layers (equivalent to 6 layers in BERT), CompFreeze was able to attain performance levels on par with full model fine-tuning. Under 50\% trainable layers, CompFreeze variants include Odd-LC, Even-LC, Upper-LC, and Lower-LC. For deeper insights, we investigate model performance when compacter modules are integrated into varying numbers of layers within BERT. This analysis provides valuable insights into determining the optimal number of compacter modules required to achieve satisfactory performance. In Figure~\ref{fig:delta_percent_comparison}, we present the F1 score for models where compacter modules are added at single, three, and six layers within BERT. Here, models are trained for 10 epochs with a learning rate of 5e-5. It is worth noting that the BERT-Base model consists of 12 layers, so one layer corresponds to approximately 8.33\%, three layers to 25\%, and six layers to 50\% of the total BERT-Base layers, respectively. Figure~\ref{fig:delta_percent_comparison}, depicts the average F1 scores computed across all choices for a particular layer. The interesting thing here is that for the DGA and APTNER datasets, all models exhibit similar performance whether compacter modules are added to one, three, or six layers. In contrast, the CyBERT model demonstrates a noticeable performance boost on the Spam dataset as the number of compacter modules increases. This implies that the optimal number of compacters depends on the specific dataset and the choice of pre-trained model. 

\begin{figure}[ht]
\centering
\begin{subfigure}[b]{0.45\textwidth}
  \centering
  \includegraphics[width=\linewidth]{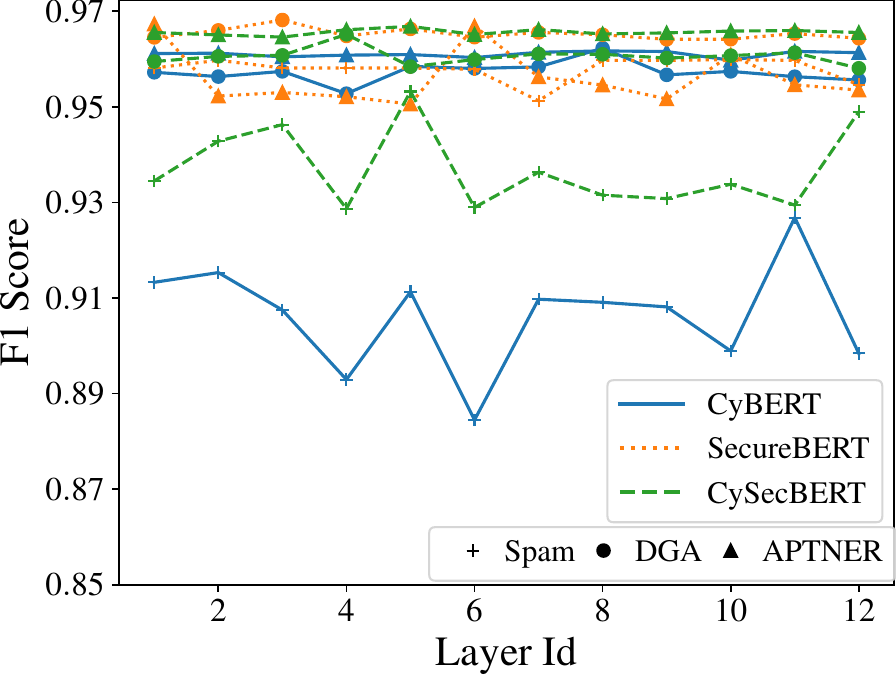}
  \caption{Compacter in single layer}
  \label{fig:delta_layer_single}
\end{subfigure}
\hfill
\begin{subfigure}[b]{0.45\textwidth}
  \centering
  \includegraphics[width=\linewidth]{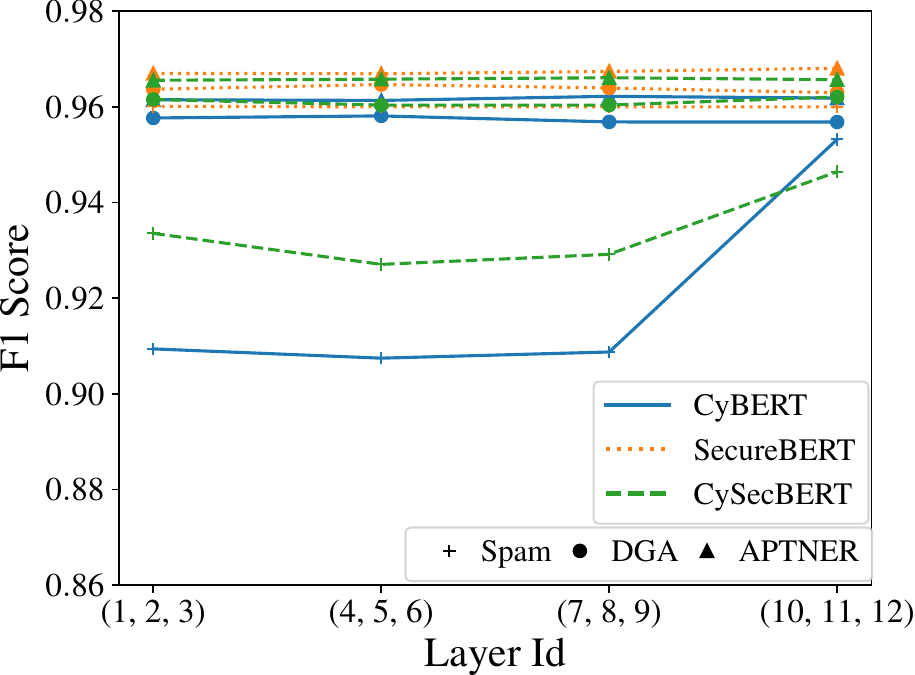}
  \caption{Compacter in three layers}
   \label{fig:delta_layer_three}
\end{subfigure}

\caption{Layer-wise model performance when adding compacter module in single and three layers.}
\label{fig:layer_f1_analysis}

\end{figure}

	
	

Drawing insights from this observation, one can begin by introducing a minimal number of trainable parameters using parameter-efficient fine-tuning techniques and gradually incorporate additional parameters to achieve satisfactory performance. For a detailed breakdown of F1 scores for each layer choice, refer to Figure~\ref{fig:layer_f1_analysis}. For single layers, we have 12 choices available, given the BERT and RoBERTa base models. In the case of three-layers, we have ${12 \choose 3}=220$ choices. However, considering that adjacent layers in a model tend to learn similar patterns, we focused on specific choices, namely \{(1,2,3), (4,5,6), (7,8,9), and (10,11,12)\}. For the six layers, we employed Odd-LC, Even-LC, Upper-LC, and Lower-LC. Notably, when compacter modules are added to a single layer, there is not a single deterministic layer that consistently achieves the best F1-score as shown in Figure~ \ref{fig:delta_layer_single}. In the case of three-layer configurations, we observe that layers (10,11,12) slightly outperform other layers for the CyBERT and CySecBERT models on the Spam dataset, as visible in Figure~\ref{fig:delta_layer_three}. However, it is essential to note that this combination does not guarantee optimal performance for all datasets and pre-trained models.  As discussed in Section \ref{subsec:lr_analysis}, the six-layer CompFreeze strategies show similar performances (as shown in Figure~\ref{fig:lr_f1}). Our results indicate that CyBERT's performance improves as the number of trainable parameters increases for the SPAM dataset. However, this trend does not hold for other datasets. In order to investigate this further, we conducted additional analysis. As a starting point, we calculated the average number of words per instance in each dataset. The average is 157 for the SPAM dataset, while for the APTNER dataset, it is 27. In contrast, the DGA dataset contains URL-type examples with significantly shorter lengths than the SPAM dataset. Before prematurely attributing this trend to text length, we conducted experiments on the IMDB (\citet{maas2011learning}) and 20NewsGroups\footnote{http://qwone.com/~jason/20Newsgroups/} datasets. These datasets have average word counts of 282 and 172 per instance, respectively. We observed behavior similar to that of the APTNER and DGA datasets. It appears that it is highly dependent on the specific task and dataset, which is a common observation in the field of machine learning.

\begin{table*}[ht]
\hspace{60pt}
\caption{Analysis of Throughput (TP) and Inference Time (IT). IT is measured in milliseconds.}
\resizebox{0.8\columnwidth}{!}{%

\begin{tabular}{cccc|cc|cc}
\hline
Task & \begin{tabular}[c]{@{}c@{}}fine-tuning\\ Strategy\end{tabular} & \multicolumn{2}{c|}{CyBERT} & \multicolumn{2}{c|}{SecureBERT} & \multicolumn{2}{c}{CySecBERT} \\ \hline
                   &              & IT    & TP     & IT    & TP     & IT    & TP     \\ \cline{3-8} 
                   & Full         & 06.57 & 517.60 & 07.55 & 517.75 & 06.59 & 517.89 \\
CTI Extraction     & Single Layer & 07.20 & 511.85 & 07.37 & 512.06 & 09.22 & 512.37 \\
                   & Six Layer    & 11.49 & 484.78 & 11.02 & 484.19 & 10.39 & 484.65 \\ \hline
                   & Full         & 10.64 & 105.59 & 07.46 & 105.45 & 08.44 & 105.61 \\
Spam Detection     & Single Layer & 08.54 & 104.76 & 07.64 & 104.62 & 09.92 & 104.77 \\
                   & Six Layer    & 11.86 & 100.77 & 10.56 & 100.49 & 11.43 & 100.78 \\ \hline 
                   & Full         & 06.70 & 517.37 & 06.87 & 518.14 & 07.68 & 518.85 \\
DGA Classification & Single Layer & 07.63 & 512.02 & 07.68 & 511.42 & 07.30 & 511.26 \\
                   & Six Layer    & 11.29 & 484.55 & 13.16 & 478.88 & 10.76 & 484.77 \\ \hline
\end{tabular}%
}

\label{tab:inference_time_table}
\end{table*}

\textbf{Inference Time and Throughput Analysis:}\label{subsec:inference_time}
CompFreeze facilitates parameter-efficient fine-tuning of pre-trained models. This in turn significantly reduces training time by training only a small number of additional parameters while keeping the PLM parameters fixed. It is important to note that during the inference stage for a given task, the total number of parameters is slightly higher than the original PLM, resulting in increased inference time. The inference time scales proportionally with the number of added parameters.

In Table~\ref{tab:inference_time_table}, we present inference time (milliseconds) and throughput when integrating compacter modules into varying layers of different cybersecurity PLMs. To calculate inference time, we individually processed 300 samples and reported the average time. For throughput calculation, we considered 100 batches of size 32 and computed the average throughput, assuming the model can handle a maximum of 32 samples simultaneously. To compute inference time and throughput properly, we perform synchronization between the host and device (i.e., GPU and CPU), so that the time is recorded only after the process running on the GPU is finished. We observed an increase in inference time ranging from 10\% to 60\% and a decrease in throughput by approximately 3\% to 6\% when comparing setups where compacter modules are added into single and six layers of PLMs. It is important to note that the exact numbers may vary depending on your hardware configuration. As demonstrated in Figure~\ref{fig:delta_percent_comparison}, the performance gained from adding more parameters to the model depends on the dataset. 
This accuracy-time trade-off analysis is particularly valuable for businesses offering AI services to a large user base where latency could be an issue.

%
	
	

\subsubsection{Analysis on Large Language Models} \label{sec:LLM_analysis}

In this section, we evaluate different LLMs with different CompFreeze-based models in terms of their accuracy. We conduct various experiments to verify the effectiveness of enhancing CompFreeze-based models with LLMs. We show that together they can be utilized for more efficient and robust cybersecurity applications.  It is important to note that for evaluation purposes, we collected a subset of samples from all the three datasets and displays the performance of models on these samples. We sampled around 208 sentences from Spam detection dataset, 200 domains and 245 sentences from DGA classification and CTI extraction (APTNER) dataset respectively. We sampled different numbers of sentences for each dataset because we wanted to be sure there is no class imbalance in these samples.

\begin{figure}[ht]
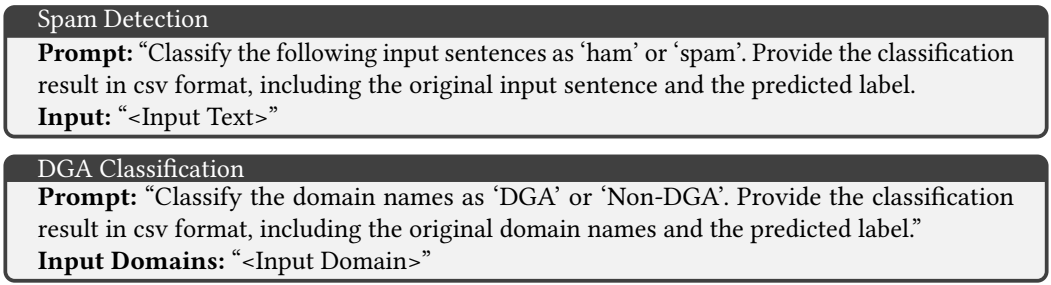

	\centering
	\begin{tcolorbox}[title=Spam Detection, boxsep=0.22pt, top=1pt, bottom=1pt]
  \textbf{Prompt:} ``Classify the following input sentences as `ham' or `spam'. Provide the classification result in csv format, including the original input sentence and the predicted label.

  \textbf{Input:} ``<Input Text>''
\end{tcolorbox}
    \begin{tcolorbox}[title=DGA Classification, boxsep=0.22pt, top=1pt, bottom=1pt]
  \textbf{Prompt:} ``Classify the domain names as `DGA' or `Non-DGA'. Provide the classification result in csv format, including the original domain names and the predicted label.''

  \textbf{Input Domains:} ``<Input Domain>''
\end{tcolorbox}

	\caption{Prompts used for Spam detection and DGA classification.}
	\label{fig:prompt_spam_dga}
\end{figure}

\begin{figure}[ht]
	\centering
	
   \begin{subfigure}[b]{\columnwidth}
	
		\begin{tcolorbox}[title=CTI Extraction, boxsep=0.22pt, top=1pt, bottom=1pt]
    \textbf{Prompt:} ``Considering the following named entities, \{list\_of\_entities\}, your task is to classify each token to one of the labels provided. For each token provided, generate a corresponding label from the list. Ensure that the output is in format: input\_token, predicted\_label''

    \textbf{Input:} ``<Input Token>''
    \end{tcolorbox}
		\caption{Simple prompt used for CTI extraction task.}
		\label{fig:prompt_ner_1}
	\end{subfigure}
	
	\begin{subfigure}[b]{\columnwidth}
	 \begin{tcolorbox}[title=CTI Extraction, boxsep=0.22pt, top=1pt, bottom=1pt]
  \textbf{System:} ``You are an expert Named Entity Recognition (NER) system that follows the task description given. You will classify named entities using the BIOES labeling format, where `B-’ indicates beginning of a multi-token entity, `I-’ indicates inside of multi-token entity and `E-’ indicates end of multi-token entity. `S-’ indicates a single token entity and `O’ refers to token not part of any entity.''

  \textbf{Task Instruction:} ``Consider the following named entities, \{list\_of\_entities\}, your task is to identify and classify each token to named entities. You will use the following description of named entities: 
    \begin{itemize}
     \item Named Entity Label: Named Entity Description
    \end{itemize}
  Classification Rules:
  \begin{enumerate}
  \item  While classifying, carefully consider the named entities definitions and labels of named entities provided.
  \item Only output the labeled entities in BIOES format, without any explanations or extra text.
  \item For tokens not belonging to any entity, mark them as O (Outside).
\end{enumerate}
    Output Format: Ensure that the output is in format: input token, predicted label

  \textbf{Input:}``<Input token>''
  \end{tcolorbox}
		\caption{Detailed prompt used for CTI extraction task.}
		\label{fig:prompt_ner_2}
	\end{subfigure}
	
	\caption{Different version of prompts used for CTI extraction task.}
	\label{fig:prompt_ner}
\end{figure}

\textbf{Prompt Analysis:} The performance of large language models varies significantly based on prompt quality and design. In this section, we evaluate the performance of GPT-4 and LLaMA model for different downstream tasks considered in this work. For comparison purpose, we evaluate two different prompts as shown in Figure \ref{fig:prompt_spam_dga} and Figure \ref{fig:prompt_ner}. The Table \ref{tab:Prompt_analysis} displays the performance of ChatGPT and LLaMA across different downstream task. The values are presented in terms of F1-score across all the downstream tasks. We observe that ChatGPT slightly outperforms the LLaMA on Spam and APTNER dataset whereas in case of DGA dataset, LLaMA provides the better performance. Upon inspection, we find that ChatGPT is able to identify the unusual patterns of spam inputs and distinguish them well as compared to LLaMA. For DGA classification task, LLaMA slightly outperforms ChatGPT with a difference of 1\%. For CTI extraction task (APTNER dataset), ChatGPT outperforms LLaMA achieving a weighted F1-score of 0.90 as compared to 0.88 of LLaMA. This observation can be attributed to the fact that both Spam detection and CTI extraction are complex tasks which require better understanding of input texts. ChatGPT has gained an edge in these tasks because it has more number of parameters leading to better contextual understanding of the text. In case of DGA classification, the inputs are domain names wherein identifying patterns is not a complex task.

\begin{table}[h]
\caption{Performance of LLMs on three downstream tasks.}
\label{tab:Prompt_analysis}
\begin{tabular}{|l|c|c|ll}
\cline{1-3}
Task               & \multicolumn{1}{l|}{ChatGPT (GPT-4)} & \multicolumn{1}{l|}{LLaMA} &  &  \\ \cline{1-3}
Spam Detection     & 0.94                                 & 0.92                       &  &  \\ \cline{1-3}
DGA Classification & 0.97                                 & 0.98                       &  &  \\ \cline{1-3}
CTI Extraction     & 0.90                                 & 0.88                       &  &  \\ \cline{1-3}
\end{tabular}
\end{table}

From Table \ref{tab:Prompt_analysis}, it can be observed for Spam detection and DGA classification task, LLMs are displaying accuracy of more than 0.95\%. The same cannot be observed for the CTI extraction task. This is because CTI extraction is a named entity recognition task which is a complex task wherein to get correct recognition of entities, models have to identify both entity boundaries and entity classes correctly. LLMs also require better understanding of this task which can be obtained through prompts. To evaluate LLMs on this task, we compared them on two different prompts on CTI extraction task as shown in Figure \ref{fig:prompt_ner}. Prompt 1 is a generic version while providing only the necessary details for the task. On the other hand, Prompt 2 is a more detailed version providing all important details of the task. Notably, it can be observed from Figure \ref{tab:accuracy_different_prompt} that the performance of both models increases with Prompt 2 as compared to Prompt 1. It indicates that more specific and detailed prompt enables the LLMs to better capture the nuances of CTI extraction task. These findings emphasize that designing prompts covering every aspect of task is crucial for obtaining high performance.

\begin{table}[h]
\caption{Performance of LLMs for CTI extraction on different prompts.}
\label{tab:accuracy_different_prompt}
\begin{tabular}{|l|c|c|ll}
\cline{1-3}
Prompts  & \multicolumn{1}{l|}{ChatGPT (GPT-4)} & \multicolumn{1}{l|}{LLaMA} &  &  \\ \cline{1-3}
Prompt 1 & 0.90                                 & 0.88                       &  &  \\ \cline{1-3}
Prompt 2 & 0.92                                 & 0.91                       &  &  \\ \cline{1-3}
\end{tabular}
\end{table}

\textbf{Low Confidence Analysis:} In this section, we investigate the performance of ChatGPT and LLaMA, along with different variants of CompFreeze-based models. We calculated the confidence score of CompFreeze-based SecureBERT model on all the three datasets and re-route those predictions to LLMs which have confidence score less than the threshold score. For all the datasets, we considered the threshold confidence score as 0.75. For the APTNER dataset, to calculate confidence score we took average across the sentences and then compared it with threshold. Table \ref{tab:low_confidence_1} highlights the F1-score of the LLMs along with CompFreeze-based models. It can be observed that CompFreeze-based models outperform LLMs for all downstream tasks. For Spam detection task, CompFreeze-based SecureBERT models display consistent performance, greater than both the ChatGPT and LLaMA models. The same observation can be noticed for other CompFreeze-based models as well. 

Upon inspection, we found that despite having superior performance metrics, these CompFreeze-based models were predicting some inputs with low confidence scores. For example, considering Spam detection task, the Upper-LC version of CompFreeze-based SecureBERT model predicted around 4 samples as low confident labels. To address this situation, we sent those samples with less confidence score than threshold to ChatGPT and instruct it to classify the samples. Table \ref{tab:low_confidence_2} displays the F1-score on each dataset before sending samples to LLMs and after sending them to LLMs. It can be clearly perceived that for Spam and DGA datasets, the overall accuracy of inputs increases after the low confident inputs are sent to LLMs. However, the same trend cannot be observed for CTI extraction task. To investigate this further, we conducted further analysis of the CTI extraction task (APTNER dataset). Upon careful examination, we found that some entities such as `URL', `File Names' are incorrectly labelled in dataset (\citet{wang2022aptner}). For example, consider an url token, it is labelled as `O' entity in the dataset and after sending this token to ChatGPT, it classified it as `URL'. It appears that this is the reason for the CTI extraction task to display a different trend as compared to other tasks. Howsoever, it is important to note that by combining LLMs with CompFreeze-based models, enterprise can create a more robust and flexible cybersecurity application with better performance gains.

\begin{table*}[ht]
\caption{Detailed comparison of LLMs with CompFreeze-based models.}
\label{tab:low_confidence_1}
\footnotesize
\begin{tabular}{c|c|c|c|c|c}
\hline
Task               & ChatGPT (GPT-4) & LLaMA & \begin{tabular}[c]{@{}c@{}}CompFreeze-based \\ CyBERT\end{tabular}                                     & \begin{tabular}[c]{@{}c@{}}CompFreeze-based\\ CysecBERT\end{tabular}                                   & \begin{tabular}[c]{@{}c@{}}CompFreeze-based\\ SecureBERT\end{tabular}                                  \\ \hline
Spam Detection     & 0.94            & 0.92  & \begin{tabular}[c]{@{}c@{}}Odd-LC: 0.95\\ Even-LC: 0.95\\ Upper-LC: 0.96\\ Lower-LC: 0.92\end{tabular} & \begin{tabular}[c]{@{}c@{}}Odd-LC: 0.93\\ Even-LC: 0.93\\ Upper-LC: 0.95\\ Lower-LC: 0.92\end{tabular} & \begin{tabular}[c]{@{}c@{}}Odd-LC: 0.98\\ Even-LC: 0.98\\ Upper-LC: 0.98\\ Lower-LC: 0.98\end{tabular} \\ \hline
DGA Classification & 0.97            & 0.98  & \begin{tabular}[c]{@{}c@{}}Odd-LC: 0.96\\ Even-LC: 0.96\\ Upper-LC: 0.96\\ Lower-LC: 0.96\end{tabular} & \begin{tabular}[c]{@{}c@{}}Odd-LC: 0.98\\ Even-LC: 0.98\\ Upper-LC: 0.98\\ Lower-LC: 0.98\end{tabular} & \begin{tabular}[c]{@{}c@{}}Odd-LC: 0.97\\ Even-LC: 0.97\\ Upper-LC: 0.97\\ Lower-LC: 0.97\end{tabular} \\ \hline
CTI Extraction     & 0.90            & 0.88  & \begin{tabular}[c]{@{}c@{}}Odd-LC: 0.96\\ Even-LC: 0.96\\ Upper-LC: 0.96\\ Lower-LC: 0.96\end{tabular} & \begin{tabular}[c]{@{}c@{}}Odd-LC: 0.92\\ Even-LC: 0.92\\ Upper-LC: 0.92\\ Lower-LC: 0.93\end{tabular} & \begin{tabular}[c]{@{}c@{}}Odd-LC: 0.96\\ Even-LC: 0.96\\ Upper-LC: 0.96\\ Lower-LC: 0.96\end{tabular} \\ \hline
\end{tabular}
\end{table*}

\begin{table}[hb]
\caption{Measures of F1-score before sending inputs to LLMs and after the predictions of LLMs. Please note, inputs here are those having confidence score less than 0.75.} 
\label{tab:low_confidence_2}
\begin{tabular}{c|cc}
\hline
Task                  & \multicolumn{2}{c}{F1-score}                         \\ \hline
\multicolumn{1}{l|}{} & \multicolumn{1}{l|}{Before LLM} & \multicolumn{1}{l}{After LLM} \\ \hline
Spam Detection        & \multicolumn{1}{c|}{0.98}       & 0.99                          \\ \hline
DGA Classification    & \multicolumn{1}{c|}{0.97}       & 0.98                          \\ \hline
CTI Extraction        & \multicolumn{1}{c|}{0.96}       & 0.96                          \\ \hline
\end{tabular}
\end{table}

\begin{table}[h]
\caption{Performance of CompFreeze-based models on `Original', `ChatGPT' and `LLaMA' labelled dataset for Spam detection task.}
\label{tab:spam_data_label}
\resizebox{0.7\columnwidth}{!}{%
\begin{tabular}{|c|c|c|c|}
\hline
Spam & \begin{tabular}[c]{@{}c@{}}Compfreeze-based\\ CyBERT\end{tabular} & \begin{tabular}[c]{@{}c@{}}Compfreeze-based\\ CysecBERT\end{tabular} & \begin{tabular}[c]{@{}c@{}}Compfreeze-based\\ SecureBERT\end{tabular} \\ \hline
Original & \begin{tabular}[c]{@{}c@{}}Odd-LC: 0.90\\ Even-LC: 0.90\\ Upper-LC: 0.88\\ Lower-LC: 0.90\end{tabular} & \begin{tabular}[c]{@{}c@{}}Odd-LC: 0.89\\ Even-LC: 0.91\\ Upper-LC: 0.91\\ Lower-LC: 0.90\end{tabular} & \begin{tabular}[c]{@{}c@{}}Odd-LC: 0.92\\ Even-LC: 0.87\\ Upper-LC: 0.90\\ Lower-LC: 0.90\end{tabular} \\ \hline
ChatGPT & \begin{tabular}[c]{@{}c@{}}Odd-LC: 0.89\\ Even-LC: 0.89\\ Upper-LC: 0.89\\ Lower-LC: 0.89\end{tabular} & \begin{tabular}[c]{@{}c@{}}Odd-LC: 0.90\\ Even-LC: 0.89\\ Upper-LC: 0.89\\ Lower-LC: 0.91\end{tabular} & \begin{tabular}[c]{@{}c@{}}Odd-LC: 0.83\\ Even-LC: 0.87\\ Upper-LC: 0.87\\ Lower-LC: 0.86\end{tabular} \\ \hline
LLaMA & \begin{tabular}[c]{@{}c@{}}Odd-LC: 0.85\\ Even-LC: 0.87\\ Upper-LC: 0.85\\ Lower-LC: 0.88\end{tabular} & \begin{tabular}[c]{@{}c@{}}Odd-LC: 0.89\\ Even-LC: 0.89\\ Upper-LC: 0.88\\ Lower-LC: 0.89\end{tabular} & \begin{tabular}[c]{@{}c@{}}Odd-LC: 0.86\\ Even-LC: 0.87\\ Upper-LC: 0.88\\ Lower-LC: 0.88\end{tabular} \\ \hline
\end{tabular}%
}
\end{table}

\begin{table}[h]
\caption{Performance of CompFreeze-based models on `Original', `ChatGPT' and `LLaMA' labelled dataset for DGA classification task.}
\label{tab:dga_data_label}
\resizebox{0.7\columnwidth}{!}{%
\begin{tabular}{|c|c|c|c|}
\hline
DGA & \begin{tabular}[c]{@{}c@{}}Compfreeze-based\\ CyBERT\end{tabular} & \begin{tabular}[c]{@{}c@{}}Compfreeze-based\\ CysecBERT\end{tabular} & \begin{tabular}[c]{@{}c@{}}Compfreeze-based\\ SecureBERT\end{tabular} \\ \hline
Original & \begin{tabular}[c]{@{}c@{}}Odd-LC: 0.77\\ Even-LC: 0.78\\ Upper-LC: 0.91\\ Lower-LC: 0.73\end{tabular} & \begin{tabular}[c]{@{}c@{}}Odd-LC: 0.94\\ Even-LC: 0.92\\ Upper-LC: 0.92\\ Lower-LC: 0.94\end{tabular} & \begin{tabular}[c]{@{}c@{}}Odd-LC: 0.92\\ Even-LC: 0.93\\ Upper-LC: 0.93\\ Lower-LC: 0.93\end{tabular} \\ \hline
ChatGPT & \begin{tabular}[c]{@{}c@{}}Odd-LC: 0.92\\ Even-LC: 0.92\\ Upper-LC: 0.92\\ Lower-LC: 0.91\end{tabular} & \begin{tabular}[c]{@{}c@{}}Odd-LC: 0.94\\ Even-LC: 0.91\\ Upper-LC: 0.94\\ Lower-LC: 0.94\end{tabular} & \begin{tabular}[c]{@{}c@{}}Odd-LC: 0.92\\ Even-LC: 0.92\\ Upper-LC: 0.93\\ Lower-LC: 0.94\end{tabular} \\ \hline
LLaMA & \begin{tabular}[c]{@{}c@{}}Odd-LC: 0.92\\ Even-LC: 0.91\\ Upper-LC: 0.91\\ Lower-LC: 0.77\end{tabular} & \begin{tabular}[c]{@{}c@{}}Odd-LC: 0.93\\ Even-LC: 0.92\\ Upper-LC: 0.94\\ Lower-LC: 0.94\end{tabular} & \begin{tabular}[c]{@{}c@{}}Odd-LC: 0.92\\ Even-LC: 0.92\\ Upper-LC: 0.92\\ Lower-LC: 0.93\end{tabular} \\ \hline
\end{tabular}%
}
\end{table}

\begin{table}[h]
\caption{Performance of CompFreeze-based models on `Original', `ChatGPT' and `LLaMA' labelled dataset for CTI extraction task.}
\label{tab:ner_data_label}
\resizebox{0,7\columnwidth}{!}{%
\begin{tabular}{|c|c|c|c|}
\hline
CTI & \begin{tabular}[c]{@{}c@{}}Compfreeze-based\\ CyBERT\end{tabular} & \begin{tabular}[c]{@{}c@{}}Compfreeze-based\\ CysecBERT\end{tabular} & \begin{tabular}[c]{@{}c@{}}Compfreeze-based\\ SecureBERT\end{tabular} \\ \hline
Original & \begin{tabular}[c]{@{}c@{}}Odd-LC: 0.97\\ Even-LC: 0.96\\ Upper-LC: 0.97\\ Lower-LC: 0.97\end{tabular} & \begin{tabular}[c]{@{}c@{}}Odd-LC: 0.94\\ Even-LC: 0.94\\ Upper-LC: 0.93\\ Lower-LC: 0.94\end{tabular} & \begin{tabular}[c]{@{}c@{}}Odd-LC: 0.96\\ Even-LC: 0.96\\ Upper-LC: 0.97\\ Lower-LC: 0.96\end{tabular} \\ \hline
ChatGPT & \begin{tabular}[c]{@{}c@{}}Odd-LC: 0.95\\ Even-LC: 0.95\\ Upper-LC: 0.95\\ Lower-LC: 0.95\end{tabular} & \begin{tabular}[c]{@{}c@{}}Odd-LC: 0.95\\ Even-LC: 0.95\\ Upper-LC: 0.94\\ Lower-LC: 0.93\end{tabular} & \begin{tabular}[c]{@{}c@{}}Odd-LC: 0.94\\ Even-LC: 0.95\\ Upper-LC: 0.94\\ Lower-LC: 0.95\end{tabular} \\ \hline
LLaMA & \begin{tabular}[c]{@{}c@{}}Odd-LC: 0.96\\ Even-LC: 0.96\\ Upper-LC: 0.96\\ Lower-LC: 0.96\end{tabular} & \begin{tabular}[c]{@{}c@{}}Odd-LC: 0.93\\ Even-LC: 0.93\\ Upper-LC: 0.93\\ Lower-LC: 0.92\end{tabular} & \begin{tabular}[c]{@{}c@{}}Odd-LC: 0.95\\ Even-LC: 0.95\\ Upper-LC: 0.94\\ Lower-LC: 0.94\end{tabular} \\ \hline
\end{tabular}%
}
\end{table}

\textbf{Data Labelling Analysis:} 
In section \ref{sec:data_labelling}, we have already discussed about the problem of scarcity of labelled dataset in cybersecurity domain and how LLMs can be utilized to address it. In this section, we evaluate the effectiveness of ChatGPT and LLaMA, as data labelling tools. We sampled inputs from all the three datasets as discussed in Section \ref{sec:LLM_analysis} and sent those samples to ChatGPT and LLaMA for labelling purposes. We fine-tuned CompFreeze-based models on LLM generated and `Original' labels of samples wherein we kept the sampled inputs as train set and remaining samples as test set. The baseline models in this case are considered as models fine-tuned on `Original' labels of samples. We display the results in terms of F1-score in Table \ref{tab:spam_data_label}, Table \ref{tab:dga_data_label} and Table \ref{tab:ner_data_label} for all the three downstream tasks. For Spam detection task, Table \ref{tab:spam_data_label} shows that CompFreeze-based models fine-tuned on ChatGPT labelled datasets has accuracies similar to `Original' labelled dataset. Odd-LC and Even-LC CompFreeze-based CyBERT models shows only a slightly less performance than the baseline models. But all the CompFreeze-based models fine-tuned on LLaMA labelled dataset shows a more significant drop in performance. A lower number of parameters in a model like LLaMA as compared to GPT models lead to less capacity for learning complex patterns, which could result in lower accuracy in its labeling process. However, it's important to note that model performance also depends on various factors, such as training data quality, architecture, and task complexity. For DGA classification task in Table  \ref{tab:dga_data_label}, similar observation is visible, wherein CompFreeze-based models fine-tuned on LLM labelled datasets show good accuracy. For this dataset also, CompFreeze-based models show slightly less consistent results when fine-tuned on LLaMA generated labels particularly for Lower-LC CompFreeze-based CyBERT model. For CTI extraction task, as shown in Table \ref{tab:ner_data_label}, all the CompFreeze-based models displays almost similar accuracy level as compared to baseline models. 

These results show that CompFreeze-based models display similar performance for LLM generated datasets when compared with baseline models. These findings demonstrate the enhanced capabilities of CompFreeze based models when combined with LLMs.



\section{Discussion}\label{sec:discusion}
The dynamic nature of data in cybersecurity domain and large number of parameters present in AI models pose significant obstacles to fine-tuning these models due to frequent updates required. As shown in Figure~\ref{fig:lr_f1}, models trained using our proposed parameter-efficient fine-tuning method perform comparably to those with full model fine-tuning. In Table~\ref{tab:parameters}, we present the percentage of trained parameters per task and the relative difference in training time for each task. Importantly, despite having fewer trainable parameters compared to vanilla PLMs, CompFreeze-based models exhibit negligible drops in model performance.

We have further evaluated the impact of learning rates on the proposed CompFreeze strategies. Depending on the chosen learning rate, the model may converge either rapidly or gradually. Bearing this in mind, we have conducted evaluations using a set of learning rates \{1e-5, 2e-5, 3e-5, 4e-5, 5e-5\} for three and ten training epochs. Across all datasets and cybersecurity PLMs, we observed the best performance for the learning rate of 5e-5. Interestingly, SecureBERT is less influenced by the choice of learning rate and demonstrates faster convergence compared to CyBERT and CySecBERT (Section~\ref{subsec:lr_analysis}). Notably, we found that when trained till convergence, all three models can achieve similar results, saving substantial time in making model selection (Section~\ref{subsec:model_comparison}). 

Initially, we proposed integrating compacters into 50\% of the BERT layers. However, we took a step further to analyze how much additional computational efficiency could be achieved while maintaining acceptable performance levels. To this end, we assess model performance when compacters are added to single or three layers of the cybersecurity PLMs. Our observations indicate that the optimal number of layers for compacter integration depends on the specific dataset and the choice of pre-trained model (Section~\ref{subsec:delta_percentage_comparison}). Finally, in Section~\ref{subsec:inference_time}, we provide an important business-oriented analysis by providing results for inference time and throughput for parameter-efficient fine-tuning. These findings can assist businesses in making informed decisions based on the accuracy-time trade-off, particularly when serving a large user base.

In this work, we propose two different strategies to augment CompFreeze-based models with LLMs. One core challenge faced by these models is the availability of labelled data for fine-tuning purposes. There is scarcity of labelled datasets in cybersecurity domain and enterprises face various issues while indulging in data labelling processes. To address this, we introduced the use of LLMs such as ChatGPT and LLaMA as data labelling tools. In Table \ref{tab:spam_data_label}, Table \ref{tab:dga_data_label} and Table \ref{tab:ner_data_label}, we show that with LLM generated data, some variants of CompFreeze-based models perform on-par with models trained on original labelled (Section \ref{sec:LLM_analysis}). Notably, these results display that LLMs can effectively label the unlabeled dataset, thereby helping enterprises which indulge in resource-intensive labelling processes. These findings suggest that LLMs can be used along with CompFreeze-based models for more reliable and flexible cybersecurity applications.

In second strategy, we evaluated LLMs as secondary mode of models where CompFreeze-based models are less confident about their predictions. Even though in Table \ref{tab:low_confidence_1}, CompFreeze-based models are out-performing LLMs with slight difference in accuracy, they still produce predictions with low confidence. We rerouted those inputs to LLMs and Table \ref{tab:low_confidence_2} displays the F1-score of datasets before and after sending those inputs to LLMs. It can be clearly observed that with LLMs in the picture, the overall accuracy is increasing for all the downstream tasks (Section \ref{sec:LLM_analysis}). To evaluate the zero-shot learning capabilities of these LLMs, we noted the accuracy of them on different datasets in Table \ref{tab:Prompt_analysis} and Table \ref{tab:accuracy_different_prompt}. Notably, these findings suggest that these LLMs are an important tool for handling low confident inputs where the CompFreeze-based models struggle. Overall, these findings suggest that with augmenting LLMs with CompFreeze-based models, enterprises can aim for a more robust cybersecurity application where less confident inputs are also handled with great accuracy (Section \ref{sec:LLM_analysis}).

Our work is focused on cybersecurity domain; however, this work can be extended into other domains, including healthcare, financial and scientific domains. Additionally, new techniques for parameter-efficient fine-tuning are emerging for large model tuning. Future works can include optimizing the integration of LLMs and CompFreeze-based models by replacing the confidence comparison technique with more robust techniques. Additionally, future studies can include integrating active learning or human-in-the-loop solution for data-labelling process to improve label quality and model adaptability.

\section{Conclusion}\label{sec:conclusion}
We proposed to use CompFreeze based models to address the challenges of full model fine-tuning in resource-constraints environment. In these models, compacter architectures are inserted into PLMs to improve their efficiency for real-world scenarios. We evaluated the model on various downstream tasks specific to cybersecurity domain and empirically demonstrated that by training only 0.06\% of total parameters and with 40\% faster training speed, CompFreeze based models perform on-par with full model fine-tuning. In this work, we explore two strategies to augment CompFreeze-based models with large language models. In the first strategy, we utilized LLMs as data-labelling tools to generate labels from unlabeled data. Furthermore, CompFreeze-based models are fine-tuned on these data, which are more robust for real-world scenarios. In second strategy, we focus on utilizing LLMs for low conference predictions. Those predictions from CompFreeze-based models having low confidence score, are re-routed to LLMs for further analysis. This strategy improves the robustness of CompFreeze-based models as these LLMs provide an auxiliary inference mechanism. In this work, we empirically demonstrate that by combining CompFreeze-based models with LLMs, we are able to achieve better model performance.

\bibliographystyle{ACM-Reference-Format}
\bibliography{peft-llm}

\end{document}